\documentclass[10pt,twocolumn,letterpaper]{article}

\usepackage{cvpr}      

\usepackage{graphicx}
\usepackage{amsmath}
\usepackage{amssymb}
\usepackage{booktabs}
\usepackage{subcaption}
\usepackage{color}
\usepackage{multirow}
\usepackage{multicol}
\usepackage{floatrow}
\usepackage{algorithm,algorithmic}
\usepackage[accsupp]{axessibility}

\usepackage[pagebackref,breaklinks,colorlinks]{hyperref}

\usepackage[capitalize]{cleveref}
\crefname{section}{Sec.}{Secs.}
\Crefname{section}{Section}{Sections}
\Crefname{table}{Table}{Tables}
\crefname{table}{Tab.}{Tabs.}

\newtheorem{thm}{Theorem}

\def \E {\mathrm{E}}

\def \D {\mathcal{D}}

\def \R {\mathbb{R}}

\def \N {\mathcal{N}}

\def \mh {\widehat{\mu}}

\def\cvprPaperID{7489} 
\def\confName{CVPR}
\def\confYear{2023}

\makeatletter
\newcommand{\specificthanks}[1]{\@fnsymbol{#1}}

\begin{document}

\title{Making Vision Transformers Efficient from A Token Sparsification View}


\author{Shuning Chang\textsuperscript{\rm 1}\thanks{Work done during an internship at Alibaba Group.}\quad Pichao Wang\textsuperscript{\rm 2}\thanks{Equal corresponding authors.}\textsuperscript{\, \specificthanks{3}}\quad Ming Lin\textsuperscript{\rm 2}\thanks{Work done at Alibaba Group, and now affiliated with Amazon.} \quad Fan Wang\textsuperscript{\rm 2}\quad David Junhao Zhang\textsuperscript{\rm 1} \\
Rong Jin\textsuperscript{\rm 2}\quad Mike Zheng Shou\textsuperscript{\rm 1}\footnotemark[2] \\ 
\textsuperscript{\rm 1}Show Lab, National University of Singapore \quad \textsuperscript{\rm 2}Alibaba Group \\
\small\{changshuning, junhao.zhang\}@u.nus.edu,\ \{fan.w, jinrong.jr\}@alibaba-inc.com,\ minglamz@amazon.com,\\
{\small\{pichaowang, mike.zheng.shou\}@gmail.com}
}

\maketitle

\begin{abstract}
The quadratic computational complexity to the number of tokens limits the practical applications of Vision Transformers (ViTs). Several works propose to prune redundant tokens to achieve efficient ViTs. However, these methods generally suffer from (i) dramatic accuracy drops, (ii) application difficulty in the local vision transformer, and (iii) non-general-purpose networks for downstream tasks. In this work, we propose a novel Semantic Token ViT (STViT), for efficient global and local vision transformers, which can also be revised to serve as backbone for downstream tasks. The semantic tokens represent cluster centers, and they are initialized by pooling image tokens in space and recovered by attention, which can adaptively represent global or local semantic information. Due to the cluster properties, a few semantic tokens can attain the same effect as vast image tokens, for both global and local vision transformers. For instance, only 16 semantic tokens on DeiT-(Tiny,Small,Base) can achieve the same accuracy with more than 100\% inference speed improvement and nearly 60\% FLOPs reduction; on Swin-(Tiny,Small,Base), we can employ 16 semantic tokens in each window to further speed it up by around 20\% with slight accuracy increase. Besides great success in image classification, we also extend our method to video recognition. In addition, we design a STViT-R(ecover) network to restore the detailed spatial information based on the STViT, making it work for downstream tasks, which is powerless for previous token sparsification methods. Experiments demonstrate that our method can achieve competitive results compared to the original networks in object detection and instance segmentation, with over 30\% FLOPs reduction for backbone. Code is available at {\url{https://github.com/changsn/STViT-R}}.
\end{abstract}

\begin{figure*}
  \begin{minipage}[t]{0.5\linewidth}
    \centering
    \includegraphics[scale=0.125]{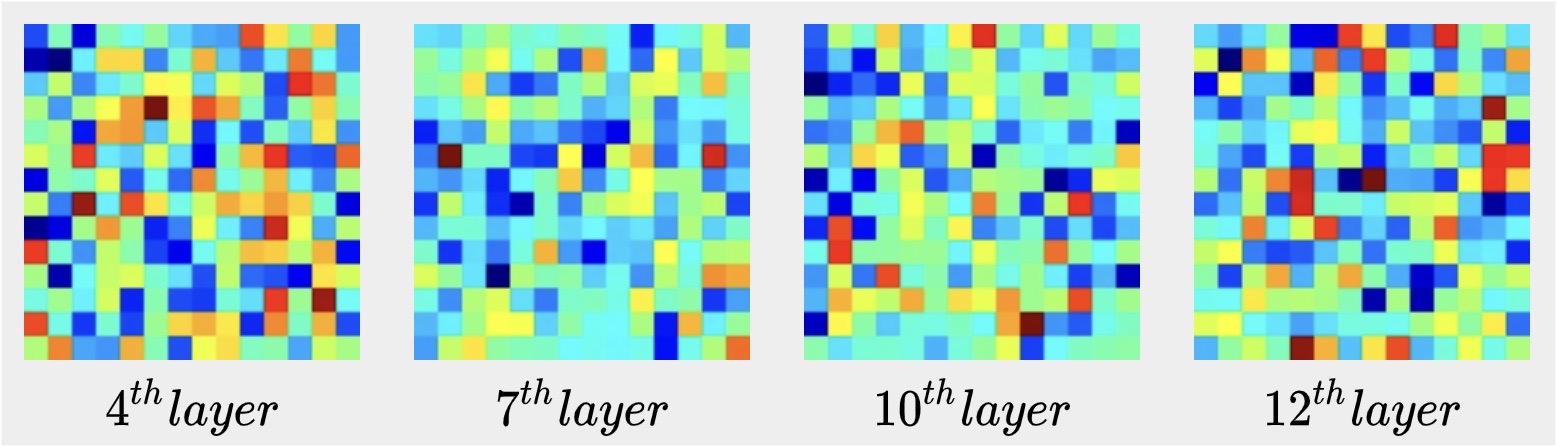}
  \end{minipage}%
  \begin{minipage}[t]{0.5\linewidth}
    \centering
    \includegraphics[scale=0.125]{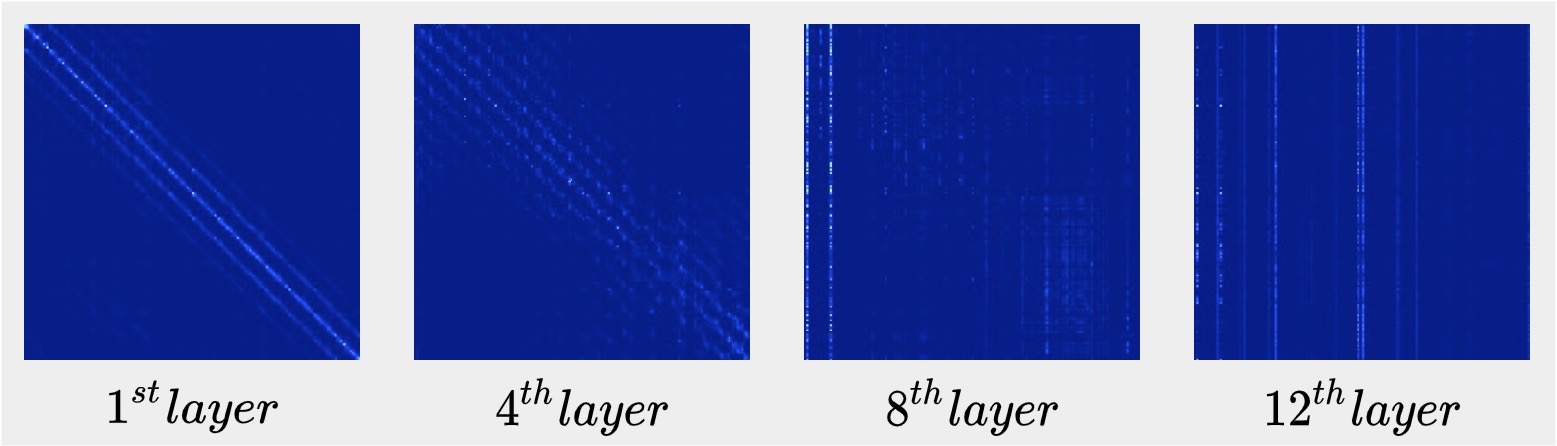}
  \end{minipage}
  \caption{Left: the attention values of class tokens (normalized and reshaped in image shape) in different self-attention layers. Right: the attention maps in different self-attention layers. Zoom-in for better visibility. \vspace{-3mm}}
  \label{fig1}
\end{figure*}

\section{Introduction}
In contrast to standard Convolutional Neural Networks (CNNs) approaches which process images pixel-by-pixel, Vision Transformers (ViTs)~\cite{dosovitskiy2020vit,touvron2021training,touvron2021going,liu2021swin,wu2021cvt} treat an image as a sequence of patch/image tokens, and have shown promising performance in prevalent visual recognition scenarios. 
However, these superior performances do not come for free: the quadratic computational complexity to the number of image tokens limits their application in practice. Previous works~\cite{song2021dynamic,zong2022self} have illustrated the large amount of redundancy in the image tokens and also shown the effect of filtering out unimportant tokens normally according to predefined scoring mechanism. However, these methods face the following challenges. Firstly, the predefined scoring mechanisms for filtering are generally imprecise. In Figure \ref{fig1}, on the left we visualize the class token values in different layers which are commonly used to score the token importance~\cite{xu2022evovit,fayyaz2021ats,liang2022evit}. Different layers have different value distributions, 
thus using these imprecise scores for filtering would lead to unsatisfactory performance. For example, EViT~\cite{liang2022evit} has an accuracy drop of 1.3\% when saving 50\% FLOPs on DeiT-S~\cite{touvron2021training}. Secondly, the remaining tokens do not distribute evenly in space any more, making them hard to work in local vision transformers\footnote{In this paper, we define the vision transformer with global self-attention (like DeiT) as global vision transformer and the vision transformer with local self-attention (like Swin) as local vision transformer.}. Finally, large-scale token pruning tremendously damages the spatial structure and positional information, and causes difficulties when applied to downstream tasks, which they do not propose a solution to deal with.

To solve these problems, we propose Semantic Token ViT (STViT), for efficient global and local vision transformers, which also can be revised to serve as backbone for downstream tasks. The proposed approach is based on the following observations: (i) unlike local CNNs which learn spatial structure of images, vision transformer discretizes feature map as tokens for global feature exploration, relieving the requirements for maintaining the whole image structure and information;
(ii) discrete tokens are more beneficial for optimization~\cite{wang2021scaled}; (iii)
in Figure~\ref{fig1}, on the right shows the attention maps in different transformer layers, and there are only several vertical lines in the deep layers, which means that only a few tokens with global semantic information matter. Thus, we argue that it is not necessary to maintain massive structured tokens for ViTs, especially in the deep layers. Employing a few discrete tokens with high-level semantic information can potentially achieve both high performance and efficiency.

In STViT, the semantic tokens represent the cluster centers, and the number of them is far less than the original image tokens, significantly reducing the computational cost. Inspired by the fact that multi-head attention can conduct the cluster center recovery (Appendix~\ref{justification}),
we only employ the off-the-shelf self-attention to generate the semantic tokens.
Specifically, the first few transformer layers are kept unchanged to obtain the image tokens with low-level features. The image tokens are then fed into our semantic token generation module (STGM) consisting of at least two transformer layers to generate semantic tokens. In each self-attention layer, the semantic tokens are input as queries, and the image tokens are fed as keys and values. The semantic tokens dynamically aggregate image tokens through the attention layers to recover cluster centers. 
In the first attention layer, the semantic tokens are initialized by an intra and inter-window spatial pooling which takes into account incorporating semantic information in each window and maximizing distance between adjacent windows.
Thanks to this spatial initialization, the semantic tokens mainly incorporate local semantic information and achieve discrete and uniform distribution in space. In the following attention layer, besides further clustering, the semantic tokens are equipped with global cluster centers, and the network can adaptively select partial semantic tokens to focus on global semantic information. After the STGM, the original image tokens are discarded, and only semantic tokens are kept for the subsequent transformer layers. Because the generation of semantic tokens is flexible and space-aware, our method can be plugged into both global and local vision transformers. The semantic tokens can be produced in each window for the local vision transformer.

Another property of STViT is its capability to serve as a backbone for downstream tasks, such as object detection and instance segmentation. Discussions have been missing in all previous methods~\cite{xu2022evovit,fayyaz2021ats,liang2022evit,ryoo2021tokenlearner,zong2022self} about how to use them in downstream task under the massive loss of spatial information during the token sparsification process, which actually seriously impedes the application of their method.
Instead, we design a novel STViT-R network based on STViT where a recovery module and dumbbell unit are adopted to periodically restore the full resolution feature map while the intermediate transformer layers continue to use semantic tokens to save computation cost, making our method work in downstream task.

The effectiveness of the proposed method is validated via a comprehensive empirical study on image and video ViT models. Only 16 semantic tokens 
on DeiT-(Tiny, Small, Base) achieve nearly 50\% inference time reduction without any accuracy degradation; on Swin-(Tiny, Small, Base), we also improve the inference throughput by nearly 20\% with slight accuracy increase. 
Moreover, the proposed STViT-R achieves promising results on object detection and instance segmentation. 
To the best of our knowledge, this is one of first works to apply the token sparsification algorithm in local vision transformers, and use the ViTs as backbones in downstream tasks after large-scale token pruning. 
Our findings in ViTs uncover that maintaining the full-size feature map is unnecessary, and a few tokens with high-level semantic representations can achieve both high performance and efficiency. Thanks to its simplicity and general-purpose ability, our method can also serve as a new efficient ViT baseline architecture and a starting point for further research from the token sparsification perspective.

\section{Related work}
\paragraph{Vision transformers.}
Vision Transformer (ViT)~\cite{dosovitskiy2020vit} first introduces a pure Transformer backbone for image classification. ViT variants further inspire the applications of transformer to various vision tasks beyond image/video classification~\cite{Yuan_2021_ICCV,touvron2021training,Xu_2021_ICCV,wang2021pyramid,wu2021cvt,touvron2021going,jiang2021all,bertasius2021space,arnab2021vivit,zhang2022morphmlp}, such as object detection~\cite{carion2020end,zhu2020deformable,zheng2020end,dai2021up}, semantic segmentation~\cite{wang2021max,wang2021end,zheng2021rethinking}, and self-supervised learning\cite{chen2021mocov3,caron2021emerging,li2021efficient}. Vanilla transformers have high computational and memory costs because the multi-head self-attention has quadratic computational complexity to the number of image tokens. Recently, various efficient ViTs have been proposed to alleviate this issue. The existing methods mainly focus on reducing the complexity of self-attention or reducing the number of tokens.  Swin Transformer~\cite{liu2021swin} adopts local self-attention, \ie, attending neighboring tokens within a constant window size, achieving a linear computational complexity in the self-attention with high performance. Many subsequent works~\cite{dong2021cswin,yu2022boat,wang2021crossformer,yang2021focal,huang2021shuffle,zhou2021elsa,chu2021twins} 
follow the local self-attention design to develop variants. Token sparsification~\cite{rao2021dynamicvit,wang2021not,pan2021ia,ryoo2021tokenlearner,chen2021chasing,song2021dynamic,yin2022vit,meng2021adavit,fayyaz2021ats,xu2022evovit,zong2022self,tang2021patch,kong2021spvit,liang2022evit} also attracts increasing attention.
\paragraph{Token sparsification.}
Token sparsification methods can be mainly categorized into hard pruning~\cite{rao2021dynamicvit,pan2021ia,chen2021chasing,song2021dynamic,yin2022vit,meng2021adavit,fayyaz2021ats,xu2022evovit,kong2021spvit,liang2022evit,tang2021patch} and soft pruning~\cite{ryoo2021tokenlearner,zong2022self}. Hard pruning methods filter out some unimportant tokens according to a predefined scoring mechanism. DynamicViT~\cite{rao2021dynamicvit}, SPViT~\cite{kong2021spvit}, and AdaViT~\cite{meng2021adavit} introduce additional prediction networks to score the tokens. Evo-ViT\cite{xu2022evovit}, ATS~\cite{fayyaz2021ats}, and EViT~\cite{liang2022evit} utilize the values of class tokens to evaluate the importance of tokens. However, it is difficult to achieve precise scoring as shown in the left of Figure~\ref{fig1}. Therefore, they usually suffer from a significant accuracy drop. For instance, EViT~\cite{liang2022evit} has an accuracy drop of 1.3\% when saving 50\% FLOPs on DeiT-S. Soft pruning methods generate new tokens from image tokens by importing additional attention networks. TokenLearner~\cite{ryoo2021tokenlearner} also argue for a few tokens to replace image tokens. However, its price is a 1.8\% accuracy drop when reducing 44\% FLOPs, which is far inferior to concurrent works. Besides performance degradation, previous methods also have the following disadvantages. First, whether or how to extend the methods to local vision transformers remains unexplored. Second, it has not been discussed about how to serve the downstream tasks like object detection and instance segmentation after the tokens are pruned.

In our method, we apply the off-the-shelf transformer layers to reduce token number. ~\cite{ma2021luna,jaegle2021perceiver,bai2021visual,zhang2019latentgnn,li2019expectation,chen2019graph} adopt similar approaches to achieve efficient non-local relationships. Our method is different from them as below: (i) our method extracts local semantic information instead of non-local relationships; (ii) the semantic tokens are a few cluster centers, which can replace the massive image tokens to achieve image classification; (iii) our method specializes in pruning tokens.
\vspace{-1mm}
\section{Method}
\vspace{-1mm}
The proposed STViT is presented in this section, which aims to construct an efficient and high-performance ViT. STViT is first introduced in Section~\ref{3.1}, followed by how to apply STViT in the local vision transformer in Section~\ref{3.2}. Based on STViT, STViT-R is developed to restore the spatial resolution for downstream tasks in Section~\ref{3.3}. 

\subsection{STViT}
\label{3.1}
\begin{figure*}[t]

    \begin{subfigure}{.55\textwidth}
        \centering
        \includegraphics[width=3in]{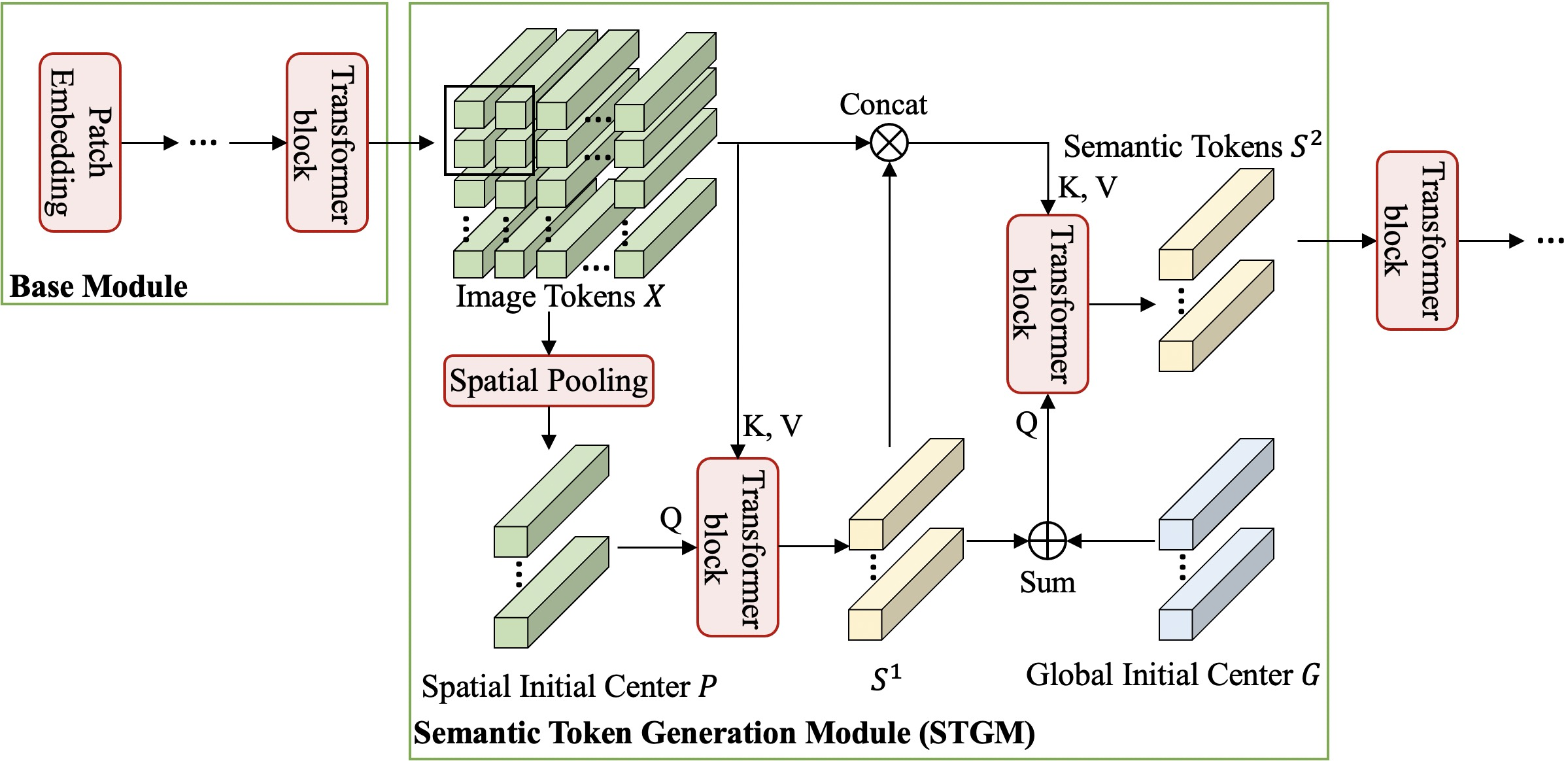}
        \caption{STViT}
        \label{arch a}
    \end{subfigure}
    \begin{subfigure}{.44\textwidth}
        \centering
        \includegraphics[width=2in]{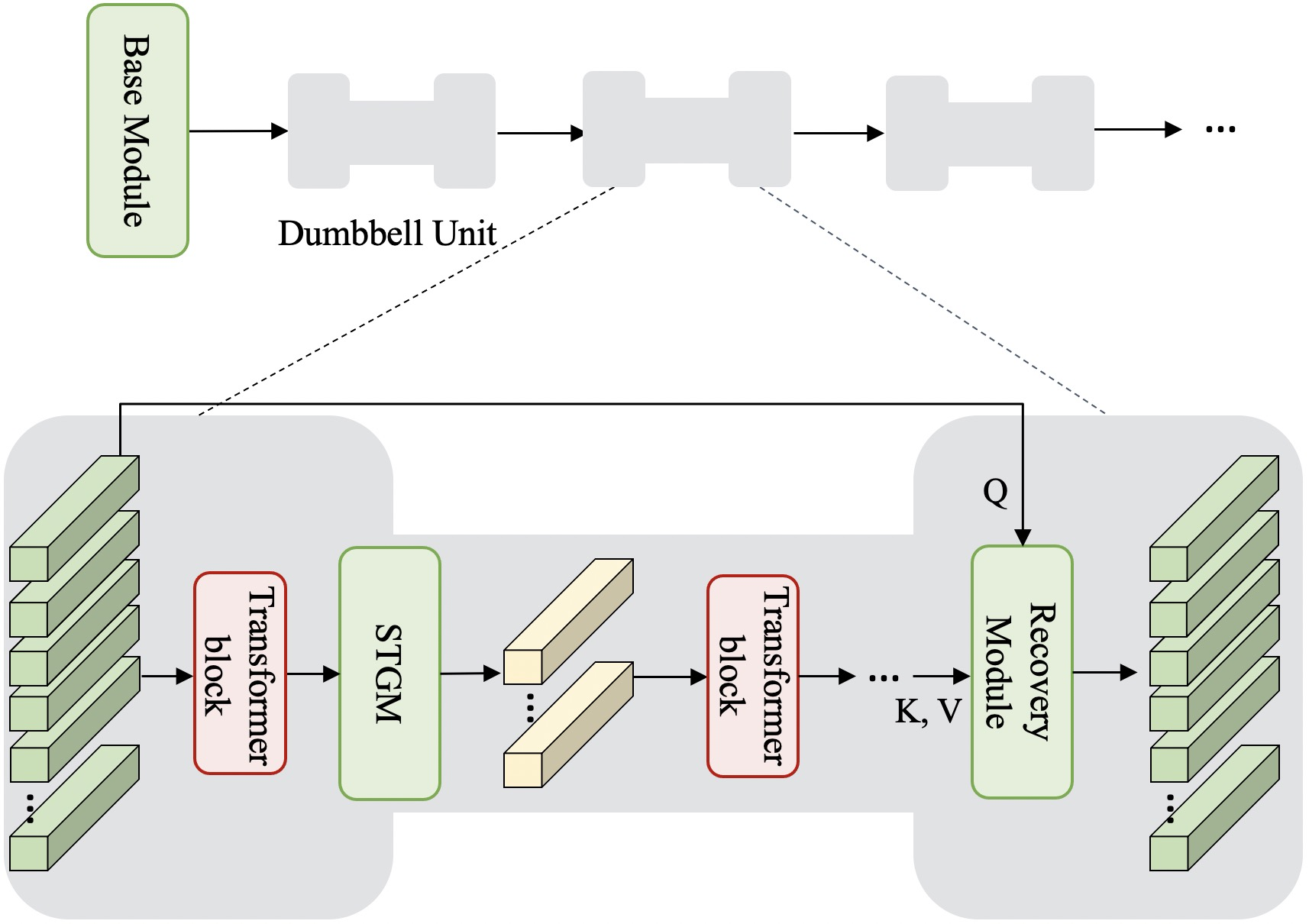}
        \caption{STViT-R}
        \label{arch b}
    \end{subfigure}
\vspace{-3mm}
\caption{The architectures of our STViT and STViT-R.\vspace{-5mm}}
\end{figure*}

\paragraph{Overall architecture.}An overview of STViT architecture is presented in Figure~\ref{arch a}. The patch embedding layer and shallow transformer layers are kept unchanged as a base module in our method. The base module copes with all the image tokens $X\in \mathbb{R}^{N_i\times C}$ to extract low-level features, where $N_i$ is the number of image tokens and $C$ is the number of channels. The image tokens are fed into the semantic token generation module (STGM) to generate $N_s$ semantic tokens $S\in \mathbb{R}^{N_s\times C}$. After the STGM, the image tokens $X$ can be discarded, and only semantic tokens $S$ with high-level semantic information are used in all the subsequent transformers. Due to $N_s \ll N_i$, our method can significantly reduce the computational cost.

\paragraph{Semantic token generation module (STGM).}
The whole image is represented by a few tokens with high-level semantic information through clustering.
Inspired by the fact that self-attention can conduct cluster center recovery (Appendix~\ref{justification}), we adopt the off-the-shelf self-attention layers to produce the semantic tokens. The STGM consists of at least two transformer layers.

The initial cluster centers $P\in \mathbb{R}^{N_s\times C}$ are generated by an spatial pooling which pools the image tokens into fixed $w_s\times w_s$ tokens, with $N_s = w_s \times w_s$. $w_s$ is generally set as $4$. The spatial pooling can be achieved by a non-parameterized adaptive spatial pooling or a super lightweight network with higher performance, intra and inter-window spatial pooling. The intentions of spatial pooling initialization are three folds. First, the initial cluster centers can distribute uniformly in space, making the generated semantic tokens more discrete and preventing the semantic tokens from collapsing to one point in the following layers. Second, the semantic tokens can be forced to represent more local and distinguished features. Finally, the representation of semantic tokens is associated with the specific spatial locations, which is the basis to allow our method to be applied in local self-attention and downstream tasks. The initial cluster centers $P$ then dynamically integrate the image tokens $X$ according to semantic information by attention mechanism. In the first transformer layer, the processing of the generation of semantic tokens can be written as
\begin{equation}
\begin{gathered}
\label{eq1}
\resizebox{0.85\hsize}{!}{
    $\hat{S^1} = MHA(P, X, X) + P, \quad
    S^1 = FFN(\hat{S^1}) + \hat{S^1}$, }
\end{gathered}
\end{equation}
where $MHA$ and $FFN$ are short for multi-head attention layer and feed-forward network, respectively, and the triplet input of $MHA$ are queries, keys, and values in turn. All the norm layers are omitted in all the equations for brevity. The initial cluster centers are produced in a window by an adaptive spatial pooling layer or an intra and inter-window spatial pooling, while the semantic tokens are generated from a global receptive field by a dynamic attention layer to ensure that they can extract high-level semantic representation. In order to further strengthen the clustering effect, we use the second transformer layer to repeat the clustering operation and guide the semantic tokens to extract global information. In this transformer layer, the semantic tokens are updated as:
\begin{equation}
\begin{gathered}
\label{eq2}
\resizebox{0.87\hsize}{!}{
    $\hat{S^2} = MHA(S^1+G, Concat(S^1, X), Concat(S^1, X)) + S^1$,}\\
    \resizebox{0.35\hsize}{!}{$S^2 = FFN(\hat{S^2}) + \hat{S^2}$},
\end{gathered}
\end{equation}
where $G\in \mathbb{R}^{N_s\times C}$ is the global cluster centers initialized by Gaussian noise and $Concat(\cdot)$ is a concatenation operation. The global cluster centers $G$ are responsible for global semantic information extraction like the class token. 
Although $S^1$ and $G$ are summed together as the queries, they can be decoupled in the attention computation as:
\begin{equation}
\begin{gathered}
    \resizebox{0.87\hsize}{!}{$A_s = (S^1\cdot W_q)\cdot((S^1+X)\cdot W_k), \quad A_g = (G\cdot W_q)\cdot((S^1+X)\cdot W_k),$}\\
    \resizebox{0.35\hsize}{!}{$A = Softmax(A_s + A_g),$}
\end{gathered}
\end{equation}
where $W_q$ and $W_k$ are the linear projection weights of queries and keys. We can see that $S^1$ and $G$ integrate the keys to generate $A_s$ and $A_g$ independently and just share a Softmax operation to produce the final attention map $A$. Therefore, besides spatial semantic information, the semantic tokens also incorporate global semantic information with a negligible additional overhead. Though very similar, the global cluster centers are actually different from the learned positional encoding. We do not add the global cluster centers to keys. Moreover, it will be shown in the experiments that 
inserting the actual learned positional encoding will cause an accuracy drop. The number of transformer layers in STGM is flexible. More transformer layers can be associated for clustering. Two transformer layers are employed by default, \ie, $S^2$ is the output of the STGM. Note that the image tokens are not updated in the STGM.

\paragraph{Intra and inter-window spatial pooling.}
Give $H\times W$ feature map $X$, we generate $N_s$($w_s\times w_s$) initial cluster centers. We uniformly split the $X$ into $w_s\times w_s$ windows $[X_w^i]^{N_s}_{i=1}$ with size $\frac{H}{w_s} \times \frac{W}{w_s}$ and each window generates one initial cluster center. To represent abundant semantic information, we take into account intra and inter-window relations, i.e., integrating important tokens in the window and maximizing the distance among  initial cluster centers in different windows. Specifically, we formulate an intra-window function $f_{intra}$ to produce a mask, $M_i=f_{intra}(X_w^i)$, which projects the input window from $\mathbb{R}^{\frac{H}{w_s} \times \frac{W}{w_s}\times C}\rightarrow\mathbb{R}^{\frac{H}{w_s} \times \frac{W}{w_s}}$. The idea is to let the intra-window function $f_{intra}$ adaptively select a combination of informative tokens in $X_w^i$, which is implemented by
\begin{equation}
\label{intra}
\vspace{-0.5mm}
\resizebox{0.9\hsize}{!}{
$M_i = Conv(GeLU(LayerNorm(DepthwiseConv(X_w^i))))$.}
\vspace{-0.5mm}
\end{equation}
Then, we compute each integrated token $\hat{P_i}=Softmax(M_i)\cdot X_w^i$ from each window, and arrange $[\hat{P_i}]^{N_s}_{i=1}$ by spatial structure to form the 2D tensor $\hat{P}\in \mathbb{R}^{w_s\times w_s \times C}$. We adopt an inter-window function $f_{inter}$ to compute the inter-window relations and generate the offset, $O=f_{inter}(\hat{P})$,to revise the mask $M$. The implementation of $f_{inter}$ is similar to Eq. \ref{intra}, except the mapping input from $\mathbb{R}^{w_s \times w_s\times C}\rightarrow\mathbb{R}^{w_s \times w_s \times \frac{HW}{w_s^2}}$, where $\frac{HW}{w_s^2}$ is the number of tokens in each window. For each window, the corresponding $O_i\in \mathbb{R}^\frac{HW}{w_s^2}$ is sliced from $O$ and reshaped to $\mathbb{R}^{\frac{H}{w_s} \times \frac{W}{w_s}}$. The $O_i$ is used to revise $M_i$. The final initial cluster center $P_i$ is computed by
\begin{equation}
\vspace{-0.5mm}
P_i = Softmax(M_i + O_i)\cdot X_w.
\vspace{-0.5mm}
\end{equation}
Our $f_{intra}$ and $f_{intra}$ are super lightweight and the introducing parameters can be negligible. For example, on DeiT-T, the  parameters only increase by 0.05\%.

\subsection{STViT in local vision transformers.}
\label{3.2}
Local self-attention has been widely used in current ViT models to balance efficiency and accuracy. As the generation of semantic tokens in STViT is flexible in space, it can be naturally applied in local self-attention. Suppose each local self-attention layer contains $N_w$ windows with size $w\times w$, we initialize $w_s\times w_s$ cluster centers in each window by our intra and inter-window spatial pooling. The total number of semantic tokens $N_s$ would be $w_s\times w_s\times N_w$. $w$ and $w_s$ are set as $7$ and $3$ by default separately. As a result, our method compresses more than 80\% image tokens in local self-attention. In the STGM, although initial cluster centers are from $w\times w$ windows, we use larger windows with size $w_k\times w_k$ to obtain keys and values in Eq.~\ref{eq1} and Eq.~\ref{eq2} to mitigate the effect of limited window size. Other operations in STGM are kept the same as Section~\ref{3.1}.

In the local ViT models, each local transformer layer is normally followed by a cross-window connection layer, such as a shift window transformer layer following a local transformer layer on Swin Transformer~\cite{liu2021swin}. In our method, 
the attention is computed within $w_s\times w_s$ window in the local self-attention layer,
and the cross-window connection can be achieved by computing self-attention in a larger-size (\eg, $4\times w_s $) sliding window because of the rare number of tokens in each window. For the low-resolution input, our cross-window connection layer is equal to a global self-attention layer.

\vspace{-1mm}
\subsection{STViT for downstream tasks}
\vspace{-1mm}
\label{3.3}
Our method significantly reduces the computation cost by using a small number of semantic tokens, while its side effect is losing nearly all the detailed position information. High-level vision tasks, such as object detection and instance segmentation, are difficult to be executed on this extremely incomplete feature map. This issue also exists in previous works, which hinders the application of token sparsification methods. To solve this issue, we design a STViT-R network based on STViT to restore the original spatial resolution from the semantic tokens.

Our STViT-R shown in Figure~\ref{arch b} has two modifications compared with STViT. First, we adopt a recovery module to restore the spatial resolution from semantic tokens; second, we regroup the transformer layers and construct dumbbell units composed of our STViT-R. 

\vspace{-2mm}
\paragraph{Recovery module.} In the recovery module, only the self-attention layer is employed to restore the spatial resolution without any additional networks. The image tokens $X$ and semantic tokens $S$ are partitioned as $N_w^r$ windows of size $w^r\times w^r$ and $w_s^r \times w_s^r$, respectively. The image tokens in each window aggregate the semantic tokens in the corresponding window, which is represented as:
\begin{equation}
\begin{gathered}
    \hat{X} = MHA(X, S, S) + X,\quad
    X = FFN(\hat{X}) + \hat{X}.
\end{gathered}
\end{equation}
This is a reverse operation of the generation of semantic tokens, using high-level semantic information to boost the image tokens.

\vspace{-2mm}
\paragraph{Dumbbell unit.} The transformer layers are regrouped into multiple dumbbell units in our STViT-R. Each dumbbell unit consists of four parts. The transformers in the first part are responsible for coping with image tokens; the second part is the semantic token generation module; the transformer layers in the third part deal with semantic tokens; the last part is the recovery module.  
Take the application on Swin-S (STViT-R-Swin-S) as an example. One, two, two and one transformer layers are allocated for these four parts, respectively. In total, each dumbbell unit is composed of 6 transformer layers. We concatenate three dumbbell units in Stage 3 of Swin-S.
In each dumbbell unit, the intermediate transformer layers process semantic tokens with high-level semantic information to save computational cost, and the complete spatial resolution is recovered at the end. By repeating multiple dumbbell units, the detailed spatial information will be preserved by the network, which can not only enhance the classification but also serve downstream tasks.

\vspace{-2mm}
\section{Experiments}
STViT will first be applied in two representative ViT models, DeiT~\cite{touvron2021training} and Swin~\cite{liu2021swin} for image classification and video recognition. To validate that our method is effective in downstream tasks, STViT-R is then performed on object detection and instance segmentation tasks.

\vspace{-2mm}
\subsection{Image classification}
\paragraph{Settings.} For image classification, all the models are trained on the ImageNet~\cite{deng2009imagenet} with 1.28M training images and 50K validation images from 1,000 classes. By default, the semantic token generation module (STGM) employs the $5^{th}$ and $6^{th}$ transformer layers of DeiT (with 12 layers in total), employs the $3^{th}$ and $4^{th}$ transformer layers of Stage 3 of Swin-T (with 12 layers in total), and employs the $11^{th}$ and $12^{th}$ transformer layers of Stage 3 of Swin-S and Swin-B (with 24 layers in total). The image resolution in training and inference is $224\times 224$ unless otherwise specified. The batch size is 1,024. All the models are trained from scratch for 300 epochs, and the augmentation and regularization strategies follow the original papers of DeiT and Swin. No knowledge distillation algorithms are used in our experiments. The classification is performed by applying a global average pooling layer on the output tokens of the last transformer layer, followed by a linear classifier. In evaluation, the top-1 accuracy using a single crop is reported. The FLOPs computations of this paper are measured by Fvcore\footnote{\url{https://github.com/facebookresearch/fvcore}}. Throughput is measured with the batch size of 128 on a V100 GPU. 


On Swin~\cite{liu2021swin}, the semantic tokens are generated in Stage 3, and they are not downsampled in Stage 4 due to rare semantic tokens. The patch merging layer between Stage 3 and Stage 4 is replaced with a simple linear layer to double the number of channels. $w_k$ is set as 10 and 14 for two transformer layers of STGM.

\begin{table*} [t]
\small
\begin{center}
\small
\begin{tabular}{c|c|c|cccc}
\toprule
\multirow{2}{4em}{Model} & \multirow{2}{4em}{Metrics} & \multirow{2}{2em}{Base} & \multicolumn{4}{c}{No. of semantic tokens} \\
& & & 16 & 36  & 64 & 100 \\
\midrule
\multirow{3}{*}{STViT-DeiT-T} & Top-1 Acc(\%) & 72.2 & 72.2(+0.0\%) & 72.7(+0.5) & 73.0(+0.8) & 73.2(+1.0)\\
& FLOPs(G) & 1.26 & 0.53(-58\%) & 0.60(-52\%) & 0.71(-44\%) & 0.86(-32\%) \\
& Throughput(img/s) & 2752 & 5511(+101\%) & 4769(+74\%) & 4214(+53\%) & 3551(+29\%) \\
\midrule
\multirow{3}{*}{STViT-DeiT-S} & Top-1 Acc(\%) & 79.8 & 79.8(+0.0) & 80.1(+0.3) & 80.5(+0.7) & 80.6(+0.8)\\
& FLOPs(G) & 4.58 & 1.91(-58\%) & 2.20(-52\%) & 2.62(-43\%) & 3.16(-31\%) \\
& Throughput(img/s) & 1408 & 2891(+105\%) & 2542(+80\%) & 2229(+58\%) & 1837(+30\%) \\
\midrule
\multirow{3}{*}{STViT-DeiT-B} & Top-1 Acc(\%) & 81.8 & 81.8(+0.0) & 82.2(+0.4) & 82.6(+0.8) & 82.7(+0.9)\\
& FLOPs(G) & 17.58 & 7.31(-58\%) & 8.44(-52\%) & 10.04(-43\%) & 12.13(-31\%) \\
& Throughput(img/s) & 626 & 1308(+110\%) & 1150(+85\%) & 1087(+61\%) & 826(+33\%) \\ 
\bottomrule
\end{tabular}
\end{center}
\vspace{-3mm}
\caption{Applying STViT to DeiT-T, DeiT-S, and DeiT-B. The top-1 accuracy, complexity in FLOPs, and throughput are reported for different numbers of semantic tokens.}
\label{cls deit}
\end{table*}

\begin{table*} [t]
\small
\begin{center}
\small
\begin{tabular}{c|c|c|c|ccc}
\toprule
\multirow{2}{4em}{Model} & \multirow{2}{4em}{Metrics} & \multirow{2}{2em}{Base} & Move & \multicolumn{3}{c}{No. of semantic tokens} \\
& & & STGM & 4 & 9  & 16 \\
\midrule
\multirow{3}{*}{STViT-Swin-T} & Top-1 Acc(\%) & 81.3 & 81.0(-0.3\%) & 80.8(-0.5) & 81.5(+0.2) & 81.8(+0.5\%) \\
& FLOPs(G) & 4.5 & 3.14(-30\%) & 2.99(-34\%) & 3.43(-24\%) & 4.06(-10\%) \\
& Throughput(img/s) & 878 & 1124(+29\%) & 1128(+29\%) & 1061(+22\%) & 1008(+15\%)\\
\midrule
\multirow{3}{*}{STViT-Swin-S} & Top-1 Acc(\%) & 83.0 & 82.8(-0.2\%) & 82.4(-0.6\%) & 83.0(-0.0) & 83.1(+0.1\%) \\
& FLOPs(G) & 8.7 & 5.95(-32\%) & 5.95(-32\%) & 6.53(-25\%) & 7.36(-15\%) \\
& Throughput(img/s) & 551 & 739(+35\%) & 732(+34\%) & 691(+26\%) & 657(+20\%) \\
\midrule
\multirow{3}{*}{STViT-Swin-B} & Top-1 Acc(\%) & 83.5 & 83.2(-0.3\%) & 83.0(-0.5) & 83.4(-0.1) & 83.7(+0.2\%) \\
& FLOPs(G) & 15.4 & 10.48(-32\%) & 10.48(-32\%) & 11.51(-25\%) & 12.97(-16\%) \\
& Throughput(img/s) & 415 & 558(+35\%) & 551(+33\%) & 521(+26\%) & 489(+19\%) \\
\bottomrule
\end{tabular}
\end{center}
\vspace{-3mm}
\caption{Applying STViT to Swin-T, Swin-S, and Swin-B. The top-1 accuracy, complexity in FLOPs, and throughput are reported for different numbers of semantic tokens in each window. \textit{Base} indicates the corresponding original Swin model. \textit{Move STGM} indicates changing the default position of STGM.\vspace{-3mm}}
\label{cls swin}
\end{table*}

\vspace{-2mm}
\paragraph{Results.}
One of the advantages of our method is that it can be applied to both global and local vision transformers to reduce computational complexity.
Our main results on DeiT and Swin are summarized in Table \ref{cls deit} and Table \ref{cls swin}, respectively. The results of LV-ViT~\cite{jiang2021all} are illustrated in Appendix \ref{a2} due to limited space. We report the top-1 accuracy, FLOPs, and the throughput under different numbers of semantic tokens. On DeiT, the models with 16 semantic tokens achieve the same accuracy as the DeiT models with 196 tokens and save nearly 60\% FLOPs on DeiT-T, DeiT-S, and DeiT-B. With more semantic tokens, the accuracy can consistently outperform the base models. For instance, STViT-DeiT-B with 36 semantic tokens surpasses DeiT-B by 0.4\% accuracy with 52\% FLOPs reduction. 

The local vision transformer like Swin is already an efficient architecture compared to the global vision transformer, so the reduction of FLOPs on Swin models are smaller than on DeiT models. When 9 semantic tokens are used in each window, STViT-Swin models can reduce 25\% FLOPs with negligible accuracy loss on all the model sizes. If the number of tokens is reduced to 4 in each window (16 in total), a significant accuracy drop will occur, which indicates that local vision transformers need more semantic tokens than global vision transformers. We can move the STGM towards shallow layers to attain a better complexity/accuracy trade-off. In Table~\ref{cls swin}, STGM is moved by one transformer layer on STViT-Swin-T and by two layers on STViT-Swin-S and STViT-Swin-B to save over 30\% FLOPs with only about 0.3\% accuracy drops (column of ``Move STGM"). 

STViT-R equipped with recovery modules is designed for downstream tasks, while it also can perform image classification. The corresponding results are reported in Table~\ref{cls recovery}. On both Swin-S and Swin-B, STViT-R can save 33\% FLOPs with 0.3\% accuracy drop. The hyper-parameters we used in STGM are as same as STViT-Swin.

These results demonstrate that our method achieves both effectiveness and efficiency by employing a few semantic tokens to replace original image tokens. Our method reveals that constructing the tokens with high-level semantic representation is more important than maintaining structured tokens in ViTs. 
As reflected by the throughput, our method does not have overhead of memory or deployment. Compared to the total parameters, the additional parameters introduced by intra and inter-window spatial pooling is negligible(less than 0.05\%), so we do not show them.
\vspace{-2mm}

\begin{table} [t]
\small
\begin{center}
\small
\resizebox{!}{0.8cm}{
\begin{tabular}{cccc}
\toprule
Model & Top-1 Acc(\%) & FLOPs(G) & Throughput \\
\midrule
STViT-R-Swin-S & 82.7(-0.3) & 5.83(-33\%) & 717(+30\%) \\
STViT-R-Swin-B & 83.2(-0.3) & 10.26(-33\%) & 539(+30\%) \\
\bottomrule
\end{tabular}
}
\caption{STViT-R is evaluated on Swin-S and Swin-B on ImageNet. The top-1 accuracy, complexity in FLOPs, and throughput are reported.}
\label{cls recovery}
\end{center}
\vspace{-5mm}
\end{table}

\begin{table}[t]
      \centering
      \small
        \begin{tabular}{cccc}
            \toprule
            Model & Top-1 Acc & FLOPs(G) & $\triangle$ \\
            \midrule
            \multicolumn{4}{c}{DeiT-S} \\
            \midrule
            DynamicViT~\cite{rao2021dynamicvit} & 79.3 & 2.9(-37\%) & -0.5\\
            IA-RED$^2$~\cite{pan2021ia} & 79.1 & 3.2(-30\%) & -0.7\\
            PS-ViT~\cite{tang2021patch} & 79.4 & 2.6(-43\%) & -0.4 \\
            TokenLearner~\cite{ryoo2021tokenlearner} & 76.1 & 1.9(-44\%) & -1.8\\
            DGE*~\cite{song2021dynamic} & 79.7 & 3.1 (-49\%) & -0.6 \\
            A-ViT*~\cite{yin2022vit} & 78.6 & 3.6 (-39\%) & -0.3\\
            Evo-ViT~\cite{xu2022evovit} & 79.4 & 3.0(-35\%) & -0.4 \\
            EViT~\cite{liang2022evit} & 78.5 & 2.3(-50\%) & -1.3 \\
            \textbf{STViT(Ours)} & 79.8 & 1.91(\textbf{-58\%}) & \textbf{-0.0}\\
            \bottomrule
            \multicolumn{4}{c}{DeiT-B} \\
            \midrule
            IA-RED$^2$~\cite{pan2021ia} & 80.3 & 11.8(-33\%) & -1.5\\
            DynamicViT~\cite{rao2021dynamicvit} & 81.3 & 11.2(-36\%) & -0.5\\
            PS-ViT~\cite{tang2021patch} & 81.5 & 9.8(-44\%) & -0.3 \\
            TokenLearner*~\cite{ryoo2021tokenlearner} & 83.7 & 28.7(-48\%) & -1.1\\
            Evo-ViT~\cite{xu2022evovit} & 81.3 & 10.2(-33\%) & -0.5 \\
            EViT~\cite{liang2022evit} & 80.0 & 8.7(-51\%) & -1.8 \\
            \textbf{STViT(Ours)} & 81.8 & 7.31(\textbf{-58\%}) & \textbf{-0.0} \\
            \bottomrule
        \end{tabular}
\caption{Comparisons with the state-of-the-art token sparsification methods on DeiT-S and DeiT-B. $\triangle$ shows the accuracy difference between each model and its base model.$*$: their base models are
not standard DeiT models.\vspace{-5mm}} \label{sota}
\end{table}


\begin{table*} [t]
\small
\begin{center}
\small
\begin{tabular}{c|cccc|cccc|c}
\toprule
& AP$^{b}$ & AP$^{b}_{50}$ & AP$^{b}_{75}$ & AP$^{b}_{s}$ & AP$^{m}$ & AP$^{m}_{50}$ & AP$^{m}_{75}$ & AP$^{m}_{s}$ & FLOPs(G)\\
\midrule
Swin-S & 51.8 & 70.4 & 56.3 & 35.2 & 44.7 & 67.9 & 48.5 & 28.8 & 194 \\
STViT-R-Swin-S & 51.8 & 70.6 & 56.1 & 36.7 & 44.7 & 67.8 & 48.6 & 29.0 & 134(-31\%)\\
\midrule
Swin-B & 51.9 & 70.9 & 56.5 & 35.4 & 45.0 & 68.4 & 48.7 & 28.9 & 343\\
STViT-R-Swin-B & 52.2 & 70.8 & 56.8 & 36.5 & 45.2 & 68.3 & 49.1 & 29.5 & 233(-32\%) \\
\bottomrule
\end{tabular}
\end{center}
\vspace{-3mm}
\caption{Results on COCO object detection and instance segmentation under Cascade Mask R-CNN with $3\times$ schedule. The FLOPs are measured for backbones.}
\label{detection}
\end{table*}
\vspace{-1mm}

\paragraph{Comparisons with existing token sparsification methods.}
In Table~\ref{sota}, we compare STViT with the state-of-the-art token sparsification methods on DeiT-S and DeiT-B. Due to different base models used by different methods, we adopt accuracy difference between each model and its base model $\Delta$ to evaluate them for fair. Results indicate that our method achieves the lowest accuracy drop $\Delta$ with the highest FLOPs reduction, outperforming all the state-of-the-art methods in both accuracy and efficiency significantly. 

\subsection{Video recognition}

\paragraph{Setting.}
For video recognition, we apply our STViT to Video Swin~\cite{liu2022video}. All the models are pre-trained on ImageNet-1K and trained on Kinetics-400~\cite{carreira2017quo}. We generate semantic tokens from each frame as illustrated in Section \ref{3.2}. The initialization from pre-trained models and other implementation details are as same as Video Swin~\cite{liu2022video}.

\vspace{-1mm}
\paragraph{Results.}
The results are presented in Table \ref{video}. On Swin-T and Swin-S, STViT-Swin can save about 27\% FLOPs with 0.3\% accuracy drop, which shows that our method works on video recognition.

\subsection{Applications in object detection and instance segmentation}
\paragraph{Settings.} Experiments of object detection and instance segmentation are conducted on COCO 2017~\cite{lin2014microsoft}. We evaluate STViT-R with Swin in Cascade Mask R-CNN~\cite{cai2018cascade,he2017mask} detection frameworks. The $w_s$ is set to $3$. The backbone models are pre-trained on ImageNet-1K and the pre-trained results are presented in Table~\ref{cls recovery}. All the other settings follow Swin~\cite{liu2021swin}.
\vspace{-3mm}
\paragraph{Comparison to Swin Transformer.} The performance of STViT-R-Swin using the Cascade Mask R-CNN framework with $3\times$ schedule is shown in Table~\ref{detection}. Our method achieves better performance with more than 30\% FLOPs reduction for backbone on bath object detection and instance segmentation. This validates that the recovery module and dumbbell unit can restore detailed spatial information, and the global context information integrated from the semantic tokens significantly benefits object detection. Ignoring spatial structure in the intermediate layers does not affect the object detection task, which is a meaningful fact to help design efficient object detection frameworks. Another interesting finding is that our method has a remarkable improvement for small object detection which is a challenging problem in the detection community. For instance, our STViT-R-Swin-S outperforms Swin-S by 1.5\% on $AP_s^b$. 



\begin{table}
\vspace{-2mm}
\center
\resizebox{!}{1.2cm}{
\begin{tabular}{c|c|c|c}
\toprule
Model & Top-1 Acc(\%) & FLOPs(G) & Speed\\
\midrule
Swin-T & 78.8 & 88 & 779\\
STViT-Swin-T & 78.5(-0.3) & 64.4(-27\%) & 975(+25\%)\\
\midrule
Swin-S & 80.6 & 166 & 456\\
STViT-Swin-S & 80.3(-0.3) & 120.5(-27\%) & 572(+25\%)\\
\bottomrule
\end{tabular}
}
\vspace{-2mm}
 \caption{Applying STViT to Video Swin (Swin-T and Swin-S) on Kinetics-400. All the models are pre-trained on ImageNet-1K. The views are $4\times3$. The top-1 accuracy and  complexity in FLOPs are reported. Speed is evaluated by FPS.}
 \label{video}
 \small
 \end{table}

\begin{table}
\center
\resizebox{!}{1.3cm}{
\begin{tabular}{c|c|c|c}
\toprule
Spatial & Global & Learned & Top-1 Acc(\%) \\
\midrule
\checkmark &  & & 79.4 \\
\midrule
& \checkmark & & 78.7 \\
\midrule
\checkmark & \checkmark & & 79.8 \\
\midrule
\checkmark &  & \checkmark & 79.7 \\
\bottomrule
\end{tabular}
}
\vspace{-2mm}
 \caption{Accuracy with different initialization of STViT. \textit{Spatial}, \textit{Global}, and \textit{Learned} indicate spatial initialization, global initialization, and learned positional encoding methods, respectively.}
 \label{initialization}
 \small

\end{table}


\begin{figure}
\includegraphics[width=1.\linewidth]{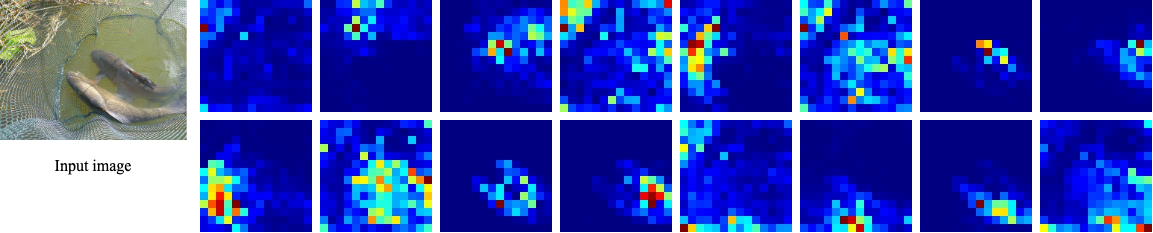}
\centering
\caption{Visualization example of attention maps in the first attention layer of STGM.\vspace{-3mm}}
\vspace{-1mm}
\label{vis}
\end{figure}

\begin{table}
\center
\small
\begin{tabular}{c|cccc}

\toprule
No. of transformers & 2 & 3 & 4 \\
\midrule
Top-1 Acc(\%) & 79.8 & 79.5 & 79.6 \\
FLOPs(G) & 1.91 & 1.97 & 2.03 \\
\bottomrule
\end{tabular}
\vspace{-2mm}
 \caption{Performance evaluation on different numbers of transformer layers in STGM. Keeping the base module containing four transformer layers unchanged.\vspace{-1mm}}
 \label{number}
 \small
\end{table}

\begin{table}
\center
\small
\begin{tabular}{c|c|c|c}
\toprule
& STViT-R & w/o DU & Reusing ST \\
\midrule
AP${^b}$ & 51.8 & 51.4 & 51.6 \\
AP${^m}$ & 44.7 & 44.4 & 44.5 \\
\bottomrule
\end{tabular}
 \caption{Ablation study on STViT-R w/o dumbbell units (\textit{w/o DU}) and reusing semantic tokens (\textit{Reusing ST)}.\vspace{-4mm}}
 \label{dumbbell units}

\end{table}

\subsection{Ablation study}
All the following ablation experiments of STViT and STViT-R are conducted on the DeiT-S and Swin-S, respectively.
\vspace{-2mm}

\paragraph{Initialization analysis.}
The semantic tokens are the cluster centers recovered by attention layers. The initialization of cluster centers induces the representation of semantic tokens. Spatial and global initialization are adopted in Section~\ref{3.1} to guide the semantic tokens to integrate local and global semantic information separately. We compare different initialization components in Table~\ref{initialization}. When performing single global initialization(3$^{rd}$ row), we replace the spatial initial cluster centers with global initial cluster centers in the first transformer layer. The accuracy of using a single initialization method is far lower than using both, which shows the effectiveness of our initialization strategy. Our global initial cluster centers look similar to learned positional encoding since both use random initialization. To verify their distinction, we experiment with a real learned positional encoding by additionally adding the global initial cluster centers to keys, which causes 0.1\% accuracy drop (the last row of Table~\ref{initialization}). Therefore, our global initialization is different from learned positional encoding.

We visualize the attention maps of the first attention layer in the STGM in Figure~\ref{vis}. 
Because the queries of this layer are spatial initial cluster centers, these attention maps visualize the local semantic information integration by attention. The attention layer groups the semantic information according to the position of initial cluster centers, which ensures to extract fine-grained semantic information and keep the difference among semantic tokens. 
The attention map in the second attention layer is visualized in Appendix~\ref{a3}, which reveals that the network fixedly selects particle semantic tokens to represent global semantic information. We also show the semantic representation of image tokens in the same transformer layer on DeiT in Appendix~\ref{a3}. Compared to original image tokens, our semantic tokens in Figure~\ref{vis} show more high-level semantic information.
\vspace{-2mm}

\paragraph{Number of transformer layers in STGM.}
Two transformer layers are adopted in the STGM by default. The effects of employing different numbers of transformer layers are shown in Table~\ref{number}. The additional  layers are from the ones behind STGM to keep the total number of layers unchanged. More transformer layers do not bring improvement.
\vspace{-2mm}

\paragraph{The effectiveness of the dumbbell unit.}
To verify the effectiveness of our dumbbell unit, we experiment STViT-R without dumbbell units, \ie, STViT equipped with only the recovery module. We employ $6^{th}$ and $7^{th}$ transformer layers to construct STGM and the last transformer layers to construct the recovery module in Stage 3. The FLOPs is the same as the full-model STViT-R for fair comparison. The results on the COCO are reported in Table~\ref{dumbbell units}. Inferior results demonstrate the effectiveness of the dumbbell unit.
\vspace{-2mm}

\paragraph{Reusing semantic tokens in dumbbell units.}
Semantic tokens are generated in each dumbbell unit. If they are produced only once in the first dumbbell unit and reused as initial cluster centers in the subsequent dumbbell units, the result is shown in Table~\ref{dumbbell units} with a slight performance drop.
\vspace{-1mm}

\section{Conclusion}
In this paper, we propose a simple and effective token sparsification method, semantic token vision transformer (STViT). Our method utilizes the clustering property of self-attention to generate a few semantic tokens with high-level information representation to replace the redundant image/video tokens, which can be applied in both global and local vision transformers. By simply configuring the recovery module, our method can be successfully applied to downstream tasks. Extensive experiments demonstrate that our method achieves better accuracy along with less inference time in most cases. The success in downstream tasks significantly boosts the development of token sparsification methods. We hope that this work can inspire more future research to pay much attention to high-level semantic representation in ViTs.

\section*{Acknowledgement}
This project is supported by the National Research Foundation, Singapore under its NRFF Award NRF-NRFF13-2021-0008, and Mike Zheng Shou's Start-Up Grant from NUS. The computational work for this article was partially performed on resources of the National Supercomputing Centre, Singapore. Shuning was supported by Alibaba Research Intern Program.


{\small

\bibliographystyle{IEEEtran}

\begin{thebibliography}{10}\itemsep=-1pt

\bibitem{arnab2021vivit}
Anurag Arnab, Mostafa Dehghani, Georg Heigold, Chen Sun, Mario Lucic, and
  Cordelia Schmid.
\newblock Vivit: A video vision transformer.
\newblock In {\em Proceedings of the IEEE/CVF international conference on
  computer vision}, pages 6836--6846, 2021.

\bibitem{bai2021visual}
Song Bai, Philip Torr, et~al.
\newblock Visual parser: Representing part-whole hierarchies with transformers.
\newblock {\em arXiv preprint arXiv:2107.05790}, 2021.

\bibitem{bertasius2021space}
Gedas Bertasius, Heng Wang, and Lorenzo Torresani.
\newblock Is space-time attention all you need for video understanding?
\newblock In {\em ICML}, volume~2, page~4, 2021.

\bibitem{cai2018cascade}
Zhaowei Cai and Nuno Vasconcelos.
\newblock Cascade r-cnn: Delving into high quality object detection.
\newblock In {\em Proceedings of the IEEE conference on computer vision and
  pattern recognition}, pages 6154--6162, 2018.

\bibitem{carion2020end}
Nicolas Carion, Francisco Massa, Gabriel Synnaeve, Nicolas Usunier, Alexander
  Kirillov, and Sergey Zagoruyko.
\newblock End-to-end object detection with transformers.
\newblock In {\em European conference on computer vision}, pages 213--229.
  Springer, 2020.

\bibitem{caron2021emerging}
Mathilde Caron, Hugo Touvron, Ishan Misra, Herv{\'e} J{\'e}gou, Julien Mairal,
  Piotr Bojanowski, and Armand Joulin.
\newblock Emerging properties in self-supervised vision transformers.
\newblock In {\em Proceedings of the IEEE/CVF International Conference on
  Computer Vision}, pages 9650--9660, 2021.

\bibitem{carreira2017quo}
Joao Carreira and Andrew Zisserman.
\newblock Quo vadis, action recognition? a new model and the kinetics dataset.
\newblock In {\em proceedings of the IEEE Conference on Computer Vision and
  Pattern Recognition}, pages 6299--6308, 2017.

\bibitem{chen2021chasing}
Tianlong Chen, Yu Cheng, Zhe Gan, Lu Yuan, Lei Zhang, and Zhangyang Wang.
\newblock Chasing sparsity in vision transformers: An end-to-end exploration.
\newblock {\em Advances in Neural Information Processing Systems}, 34, 2021.

\bibitem{chen2021mocov3}
Xinlei Chen*, Saining Xie*, and Kaiming He.
\newblock An empirical study of training self-supervised vision transformers.
\newblock {\em arXiv preprint arXiv:2104.02057}, 2021.

\bibitem{chen2019graph}
Yunpeng Chen, Marcus Rohrbach, Zhicheng Yan, Yan Shuicheng, Jiashi Feng, and
  Yannis Kalantidis.
\newblock Graph-based global reasoning networks.
\newblock In {\em Proceedings of the IEEE/CVF Conference on Computer Vision and
  Pattern Recognition}, pages 433--442, 2019.

\bibitem{chu2021twins}
Xiangxiang Chu, Zhi Tian, Yuqing Wang, Bo Zhang, Haibing Ren, Xiaolin Wei,
  Huaxia Xia, and Chunhua Shen.
\newblock Twins: Revisiting the design of spatial attention in vision
  transformers.
\newblock {\em Advances in Neural Information Processing Systems}, 34, 2021.

\bibitem{dai2021up}
Zhigang Dai, Bolun Cai, Yugeng Lin, and Junying Chen.
\newblock Up-detr: Unsupervised pre-training for object detection with
  transformers.
\newblock In {\em Proceedings of the IEEE/CVF Conference on Computer Vision and
  Pattern Recognition}, pages 1601--1610, 2021.

\bibitem{deng2009imagenet}
Jia Deng, Wei Dong, Richard Socher, Li-Jia Li, Kai Li, and Li Fei-Fei.
\newblock Imagenet: A large-scale hierarchical image database.
\newblock In {\em 2009 IEEE conference on computer vision and pattern
  recognition}, pages 248--255. Ieee, 2009.

\bibitem{dong2021cswin}
Xiaoyi Dong, Jianmin Bao, Dongdong Chen, Weiming Zhang, Nenghai Yu, Lu Yuan,
  Dong Chen, and Baining Guo.
\newblock Cswin transformer: A general vision transformer backbone with
  cross-shaped windows.
\newblock {\em arXiv preprint arXiv:2107.00652}, 2021.

\bibitem{dosovitskiy2020vit}
Alexey Dosovitskiy, Lucas Beyer, Alexander Kolesnikov, Dirk Weissenborn,
  Xiaohua Zhai, Thomas Unterthiner, Mostafa Dehghani, Matthias Minderer, Georg
  Heigold, Sylvain Gelly, Jakob Uszkoreit, and Neil Houlsby.
\newblock An image is worth 16x16 words: Transformers for image recognition at
  scale.
\newblock {\em ICLR}, 2021.

\bibitem{fayyaz2021ats}
Mohsen Fayyaz, Soroush~Abbasi Koohpayegani, Farnoush~Rezaei Jafari, Sunando
  Sengupta, Hamid Reza~Vaezi Joze, Eric Sommerlade, Hamed Pirsiavash, and
  Juergen Gall.
\newblock Adaptive token sampling for efficient vision transformers.
\newblock In {\em European Conference on Computer Vision (ECCV)}, 2022.

\bibitem{he2017mask}
Kaiming He, Georgia Gkioxari, Piotr Doll{\'a}r, and Ross Girshick.
\newblock Mask r-cnn.
\newblock In {\em Proceedings of the IEEE international conference on computer
  vision}, pages 2961--2969, 2017.

\bibitem{huang2021shuffle}
Zilong Huang, Youcheng Ben, Guozhong Luo, Pei Cheng, Gang Yu, and Bin Fu.
\newblock Shuffle transformer: Rethinking spatial shuffle for vision
  transformer.
\newblock {\em arXiv preprint arXiv:2106.03650}, 2021.

\bibitem{jaegle2021perceiver}
Andrew Jaegle, Felix Gimeno, Andy Brock, Oriol Vinyals, Andrew Zisserman, and
  Joao Carreira.
\newblock Perceiver: General perception with iterative attention.
\newblock In {\em International conference on machine learning}, pages
  4651--4664. PMLR, 2021.

\bibitem{jiang2021all}
Zi-Hang Jiang, Qibin Hou, Li Yuan, Daquan Zhou, Yujun Shi, Xiaojie Jin, Anran
  Wang, and Jiashi Feng.
\newblock All tokens matter: Token labeling for training better vision
  transformers.
\newblock {\em Advances in Neural Information Processing Systems}, 34, 2021.

\bibitem{kong2021spvit}
Zhenglun Kong, Peiyan Dong, Xiaolong Ma, Xin Meng, Wei Niu, Mengshu Sun, Bin
  Ren, Minghai Qin, Hao Tang, and Yanzhi Wang.
\newblock Spvit: Enabling faster vision transformers via soft token pruning.
\newblock {\em arXiv preprint arXiv:2112.13890}, 2021.

\bibitem{li2021efficient}
Chunyuan Li, Jianwei Yang, Pengchuan Zhang, Mei Gao, Bin Xiao, Xiyang Dai, Lu
  Yuan, and Jianfeng Gao.
\newblock Efficient self-supervised vision transformers for representation
  learning.
\newblock {\em arXiv preprint arXiv:2106.09785}, 2021.

\bibitem{li2019expectation}
Xia Li, Zhisheng Zhong, Jianlong Wu, Yibo Yang, Zhouchen Lin, and Hong Liu.
\newblock Expectation-maximization attention networks for semantic
  segmentation.
\newblock In {\em Proceedings of the IEEE/CVF International Conference on
  Computer Vision}, pages 9167--9176, 2019.

\bibitem{liang2022evit}
Youwei Liang, Chongjian Ge, Zhan Tong, Yibing Song, Jue Wang, and Pengtao Xie.
\newblock Not all patches are what you need: Expediting vision transformers via
  token reorganizations.
\newblock In {\em International Conference on Learning Representations}, 2022.

\bibitem{lin2014microsoft}
Tsung-Yi Lin, Michael Maire, Serge Belongie, James Hays, Pietro Perona, Deva
  Ramanan, Piotr Doll{\'a}r, and C~Lawrence Zitnick.
\newblock Microsoft coco: Common objects in context.
\newblock In {\em European conference on computer vision}, pages 740--755.
  Springer, 2014.

\bibitem{liu2021swin}
Ze Liu, Yutong Lin, Yue Cao, Han Hu, Yixuan Wei, Zheng Zhang, Stephen Lin, and
  Baining Guo.
\newblock Swin transformer: Hierarchical vision transformer using shifted
  windows.
\newblock In {\em Proceedings of the IEEE/CVF International Conference on
  Computer Vision}, pages 10012--10022, 2021.

\bibitem{liu2022video}
Ze Liu, Jia Ning, Yue Cao, Yixuan Wei, Zheng Zhang, Stephen Lin, and Han Hu.
\newblock Video swin transformer.
\newblock In {\em Proceedings of the IEEE/CVF Conference on Computer Vision and
  Pattern Recognition}, pages 3202--3211, 2022.

\bibitem{ma2021luna}
Xuezhe Ma, Xiang Kong, Sinong Wang, Chunting Zhou, Jonathan May, Hao Ma, and
  Luke Zettlemoyer.
\newblock Luna: Linear unified nested attention.
\newblock {\em Advances in Neural Information Processing Systems},
  34:2441--2453, 2021.

\bibitem{meng2021adavit}
Lingchen Meng, Hengduo Li, Bor-Chun Chen, Shiyi Lan, Zuxuan Wu, Yu-Gang Jiang,
  and Ser-Nam Lim.
\newblock Adavit: Adaptive vision transformers for efficient image recognition.
\newblock {\em arXiv preprint arXiv:2111.15668}, 2021.

\bibitem{pan2021ia}
Bowen Pan, Rameswar Panda, Yifan Jiang, Zhangyang Wang, Rogerio Feris, and Aude
  Oliva.
\newblock Ia-red: Interpretability-aware redundancy reduction for vision
  transformers.
\newblock {\em Advances in Neural Information Processing Systems}, 34, 2021.

\bibitem{rao2021dynamicvit}
Yongming Rao, Wenliang Zhao, Benlin Liu, Jiwen Lu, Jie Zhou, and Cho-Jui Hsieh.
\newblock Dynamicvit: Efficient vision transformers with dynamic token
  sparsification.
\newblock {\em Advances in neural information processing systems}, 34, 2021.

\bibitem{ryoo2021tokenlearner}
Michael~S. Ryoo, AJ Piergiovanni, Anurag Arnab, Mostafa Dehghani, and Anelia
  Angelova.
\newblock Tokenlearner: Adaptive space-time tokenization for videos.
\newblock In {\em Advances in Neural Information Processing Systems (NeurIPS)},
  2021.

\bibitem{song2021dynamic}
Lin Song, Songyang Zhang, Songtao Liu, Zeming Li, Xuming He, Hongbin Sun, Jian
  Sun, and Nanning Zheng.
\newblock Dynamic grained encoder for vision transformers.
\newblock In {\em Thirty-Fifth Conference on Neural Information Processing
  Systems}, 2021.

\bibitem{tang2021patch}
Yehui Tang, Kai Han, Yunhe Wang, Chang Xu, Jianyuan Guo, Chao Xu, and Dacheng
  Tao.
\newblock Patch slimming for efficient vision transformers.
\newblock {\em arXiv preprint arXiv:2106.02852}, 2021.

\bibitem{touvron2021training}
Hugo Touvron, Matthieu Cord, Matthijs Douze, Francisco Massa, Alexandre
  Sablayrolles, and Herv{\'e} J{\'e}gou.
\newblock Training data-efficient image transformers \& distillation through
  attention.
\newblock In {\em International Conference on Machine Learning}, pages
  10347--10357. PMLR, 2021.

\bibitem{touvron2021going}
Hugo Touvron, Matthieu Cord, Alexandre Sablayrolles, Gabriel Synnaeve, and
  Herv{\'e} J{\'e}gou.
\newblock Going deeper with image transformers.
\newblock In {\em Proceedings of the IEEE/CVF International Conference on
  Computer Vision}, pages 32--42, 2021.

\bibitem{wang2021max}
Huiyu Wang, Yukun Zhu, Hartwig Adam, Alan Yuille, and Liang-Chieh Chen.
\newblock Max-deeplab: End-to-end panoptic segmentation with mask transformers.
\newblock In {\em Proceedings of the IEEE/CVF Conference on Computer Vision and
  Pattern Recognition}, pages 5463--5474, 2021.

\bibitem{wang2021scaled}
Pichao Wang, Xue Wang, Hao Luo, Jingkai Zhou, Zhipeng Zhou, Fan Wang, Hao Li,
  and Rong Jin.
\newblock Scaled relu matters for training vision transformers.
\newblock In {\em Proceedings of the AAAI Conference on Artificial Intelligence
  (AAAI)}, 2022.

\bibitem{wang2021pyramid}
Wenhai Wang, Enze Xie, Xiang Li, Deng-Ping Fan, Kaitao Song, Ding Liang, Tong
  Lu, Ping Luo, and Ling Shao.
\newblock Pyramid vision transformer: A versatile backbone for dense prediction
  without convolutions.
\newblock In {\em Proceedings of the IEEE/CVF International Conference on
  Computer Vision}, pages 568--578, 2021.

\bibitem{wang2021crossformer}
Wenxiao Wang, Lu Yao, Long Chen, Binbin Lin, Deng Cai, Xiaofei He, and Wei Liu.
\newblock Crossformer: A versatile vision transformer hinging on cross-scale
  attention.
\newblock {\em arXiv preprint arXiv:2108.00154}, 2021.

\bibitem{wang2021not}
Yulin Wang, Rui Huang, Shiji Song, Zeyi Huang, and Gao Huang.
\newblock Not all images are worth 16x16 words: Dynamic transformers for
  efficient image recognition.
\newblock {\em Advances in Neural Information Processing Systems}, 34, 2021.

\bibitem{wang2021end}
Yuqing Wang, Zhaoliang Xu, Xinlong Wang, Chunhua Shen, Baoshan Cheng, Hao Shen,
  and Huaxia Xia.
\newblock End-to-end video instance segmentation with transformers.
\newblock In {\em Proceedings of the IEEE/CVF Conference on Computer Vision and
  Pattern Recognition}, pages 8741--8750, 2021.

\bibitem{wu2021cvt}
Haiping Wu, Bin Xiao, Noel Codella, Mengchen Liu, Xiyang Dai, Lu Yuan, and Lei
  Zhang.
\newblock Cvt: Introducing convolutions to vision transformers.
\newblock In {\em Proceedings of the IEEE/CVF International Conference on
  Computer Vision}, pages 22--31, 2021.

\bibitem{Xu_2021_ICCV}
Weijian Xu, Yifan Xu, Tyler Chang, and Zhuowen Tu.
\newblock Co-scale conv-attentional image transformers.
\newblock In {\em Proceedings of the IEEE/CVF International Conference on
  Computer Vision (ICCV)}, pages 9981--9990, October 2021.

\bibitem{xu2022evovit}
Yifan Xu, Zhijie Zhang, Mengdan Zhang, Kekai Sheng, Ke Li, Weiming Dong, Liqing
  Zhang, Changsheng Xu, and Xing Sun.
\newblock Evo-vit: Slow-fast token evolution for dynamic vision transformer.
\newblock In {\em Proceedings of the AAAI Conference on Artificial
  Intelligence}, 2022.

\bibitem{yang2021focal}
Jianwei Yang, Chunyuan Li, Pengchuan Zhang, Xiyang Dai, Bin Xiao, Lu Yuan, and
  Jianfeng Gao.
\newblock Focal self-attention for local-global interactions in vision
  transformers, 2021.

\bibitem{yin2022vit}
Hongxu Yin, Arash Vahdat, Jose~M Alvarez, Arun Mallya, Jan Kautz, and Pavlo
  Molchanov.
\newblock A-vit: Adaptive tokens for efficient vision transformer.
\newblock In {\em Proceedings of the IEEE/CVF Conference on Computer Vision and
  Pattern Recognition}, pages 10809--10818, 2022.

\bibitem{yu2022boat}
Tan Yu, Gangming Zhao, Ping Li, and Yizhou Yu.
\newblock Boat: Bilateral local attention vision transformer.
\newblock {\em arXiv preprint arXiv:2201.13027}, 2022.

\bibitem{Yuan_2021_ICCV}
Li Yuan, Yunpeng Chen, Tao Wang, Weihao Yu, Yujun Shi, Zi-Hang Jiang,
  Francis~E.H. Tay, Jiashi Feng, and Shuicheng Yan.
\newblock Tokens-to-token vit: Training vision transformers from scratch on
  imagenet.
\newblock In {\em Proceedings of the IEEE/CVF International Conference on
  Computer Vision (ICCV)}, pages 558--567, October 2021.

\bibitem{zhang2022morphmlp}
David~Junhao Zhang, Kunchang Li, Yali Wang, Yunpeng Chen, Shashwat Chandra, Yu
  Qiao, Luoqi Liu, and Mike~Zheng Shou.
\newblock Morphmlp: An efficient mlp-like backbone for spatial-temporal
  representation learning.
\newblock In {\em Computer Vision--ECCV 2022: 17th European Conference, Tel
  Aviv, Israel, October 23--27, 2022, Proceedings, Part XXXV}, pages 230--248.
  Springer, 2022.

\bibitem{zhang2019latentgnn}
Songyang Zhang, Xuming He, and Shipeng Yan.
\newblock Latentgnn: Learning efficient non-local relations for visual
  recognition.
\newblock In {\em International Conference on Machine Learning}, pages
  7374--7383. PMLR, 2019.

\bibitem{zheng2020end}
Minghang Zheng, Peng Gao, Renrui Zhang, Kunchang Li, Xiaogang Wang, Hongsheng
  Li, and Hao Dong.
\newblock End-to-end object detection with adaptive clustering transformer.
\newblock {\em arXiv preprint arXiv:2011.09315}, 2020.

\bibitem{zheng2021rethinking}
Sixiao Zheng, Jiachen Lu, Hengshuang Zhao, Xiatian Zhu, Zekun Luo, Yabiao Wang,
  Yanwei Fu, Jianfeng Feng, Tao Xiang, Philip~HS Torr, et~al.
\newblock Rethinking semantic segmentation from a sequence-to-sequence
  perspective with transformers.
\newblock In {\em Proceedings of the IEEE/CVF conference on computer vision and
  pattern recognition}, pages 6881--6890, 2021.

\bibitem{zhou2021elsa}
Jingkai Zhou, Pichao Wang, Fan Wang, Qiong Liu, Hao Li, and Rong Jin.
\newblock Elsa: Enhanced local self-attention for vision transformer.
\newblock {\em arXiv preprint arXiv:2112.12786}, 2021.

\bibitem{zhu2020deformable}
Xizhou Zhu, Weijie Su, Lewei Lu, Bin Li, Xiaogang Wang, and Jifeng Dai.
\newblock Deformable detr: Deformable transformers for end-to-end object
  detection.
\newblock In {\em International Conference on Learning Representations}, 2020.

\bibitem{zong2022self}
Zhuofan Zong, Kunchang Li, Guanglu Song, Yali Wang, Yu Qiao, Biao Leng, and Yu
  Liu.
\newblock Self-slimmed vision transformer.
\newblock In {\em Computer Vision--ECCV 2022: 17th European Conference, Tel
  Aviv, Israel, October 23--27, 2022, Proceedings, Part XI}, pages 432--448.
  Springer, 2022.

\end{thebibliography}
}

\clearpage
\section{Appendix}

\subsection{Computational complexity analysis}
\paragraph{Global vision transformer (DeiT).}
We define that the number of image tokens is N, the number of semantic tokens is M, and their dimension is C. The patch embedding layer is neglected. The computational complexity of a global transformer processing image tokens (IT) is:
\begin{equation}
    \begin{gathered}
    \Omega(MHA(IT)) = 4NC^2 + 2N^2C, \\
    \Omega(FFN(IT)) = 8NC^2.
    \end{gathered}
\end{equation}
The computational complexity of a global transformer processing semantic tokens (ST) is:
\begin{equation}
    \begin{gathered}
    \Omega(MHA(ST)) = 4MC^2 + 2M^2C, \\
    \Omega(FFN(ST)) = 8MC^2.
    \end{gathered}
\end{equation}
The relationships between computational complexity and token number in attention and FFN are quadratic and linear, respectively. Due to the $N\ll M$, our method significantly reduces the cost of transformers, especially the attention.
The computational complexity of STGM is:
\begin{equation}
    \begin{gathered}
    \Omega(STGM) = 2MC^2+2NC^2 + 2MNC.
    \end{gathered}
\end{equation}
In global vision transformers, our STGM is also an efficient module.
The computational complexity of whole DeiT and our STViT-DeiT are:
\begin{equation}
    \begin{gathered}
    \Omega(DeiT)=144NC^2 + 24N^2C, \\
    \Omega(STViT)=52NC^2+12M^2C+76MC^2\\+8N^2C+4MNC.
\end{gathered}
\end{equation}

\begin{table*} [t]
\small
\begin{center}
\small
\begin{tabular}{c|c|c|ccc}
\toprule
\multirow{2}{4em}{Model} & \multirow{2}{4em}{Metrics} & \multirow{2}{2em}{Base} & \multicolumn{3}{c}{No. of semantic tokens} \\
& & & 36  & 49 & 100\\
\midrule
\multirow{3}{*}{STViT-LV-ViT-S} & Top-1 Acc(\%) & 83.3 & 82.7(-0.6) & 82.8(-0.5) & 83.1(-0.2\%)\\
& FLOPs(G) & 6.6 & 3.69(-44\%) & 3.91(-41\%) & 4.62(-30\%)\\
& Throughput(img/s) & 1159 & 2073(+78\%) & 1933(+72\%) & 1592(+37\%)\\
\bottomrule
\end{tabular}
\vspace{-3mm}
\caption{Results of STViT on LV-ViT-S.}
\label{cls lvvit}
\end{center}
\end{table*}

\paragraph{Local vision transformer (Swin).}
We define that the number of image tokens (IT) is N, the number of image tokens in each window is W, the number of semantic tokens (ST) in each window is M, and their dimension is C. We only compute the computational complexity in each transformer.
The computational complexity of a local transformer processing image tokens (IT) is:
\begin{equation}
    \begin{gathered}
    \Omega(MHA(IT)) = 4NC^2 + 2W^2NC, \\
    \Omega(FFN(IT)) = 8NC^2.
    \end{gathered}
\end{equation}
The computational complexity of a local transformer processing image tokens (ST) is:
\begin{equation}
    \begin{gathered}
    \Omega(MHA(ST)) = 4(N/W)MC^2 + 2M^2(N/W)C, \\
    \Omega(FFN(ST)) = 8MC^2.
    \end{gathered}
\end{equation}
Swin makes the computational complexity linear to token number, while our method further reduces the computational complexity.
The computational complexity of STGM is:
\begin{equation}
    \begin{gathered}
    \Omega(STGM) = 2(N/W)MC^2+2NC^2 + 2(N/W)M^2C.
    \end{gathered}
\end{equation}


\subsection{The results on LV-ViT}
\label{a2}
\paragraph{Setting.}
In LV-ViT~\cite{jiang2021all}, by default, the STGM employs the $6^{th}$ and $7^{th}$ transformer layers of LV-ViT-S (with 16 layers in total). We downsample the token labels to match the size of our semantic tokens.

\paragraph{Results.}
The main results are shown in Table \ref{cls lvvit}. Token labelling in LV-ViT is not friendly for our method. Token labelling emphasizes the importance of all the output tokens and advocates that each output token should be associated with an individual location-specific label~\cite{jiang2021all}, while our semantic tokens generated by clustering emphasize high-level semantic information. However, we still achieve good performance.
In Table \ref{sota2}, we compare our STViT with the state-of-the-art token sparsification method EViT~\cite{liang2022evit} on LV-ViT-S. Results indicate that our method outperforms it.

\subsection{Applications in semantic segmentation}
\paragraph{Settings.} ADE20K is a widely-used semantic segmentation dataset, including a broad range of 150 semantic classes. It has 25K images in total, with 20K for training, 2K for validation, and 3K for testing. UperNet in mmseg is utilized as our base framework. The $w_s$ is set to $3$. Models are trained for 240K iterations. All the other settings follow the Swin Transformer~\cite{liu2021swin}.
\vspace{-2mm}
\paragraph{Comparison to Swin Transformers.}
Table~\ref{seg} presents the results of STViT-R-Swin on semantic segmentation. With similar FLOPs reduction, the drop on mIoU is larger compared with those in object detection tasks, which shows that our method still has a gap on dense prediction compared to the full-token network.

We analyze the relatively poor performance from two views. First, the STGM strictly prunes more than 80\% tokens by attention, which remains the high-level semantic information but loses nearly all the detailed information. Semantic segmentation is a dense pixel-level classification task, and the semantic tokens are difficult to enhance the pixel-level representation. Second, our spatial pooling layer with large kernel size in STGM and self-attention layers can be regarded as low-frequency filters. STGM filters most high-frequency information, which is necessary for semantic segmentation.

\subsection{Additional visualization}
\label{a3}
\begin{figure*}[t]
\begin{subfigure}{1.\textwidth}
\includegraphics[width=1.0\linewidth]{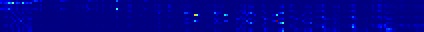}
\centering
\caption{The attention map ($16\times 216$) of the second attention layer in STGM.}
\label{app vis a}
\end{subfigure}

\begin{subfigure}{0.9\textwidth}
\includegraphics[width=1.0\linewidth]{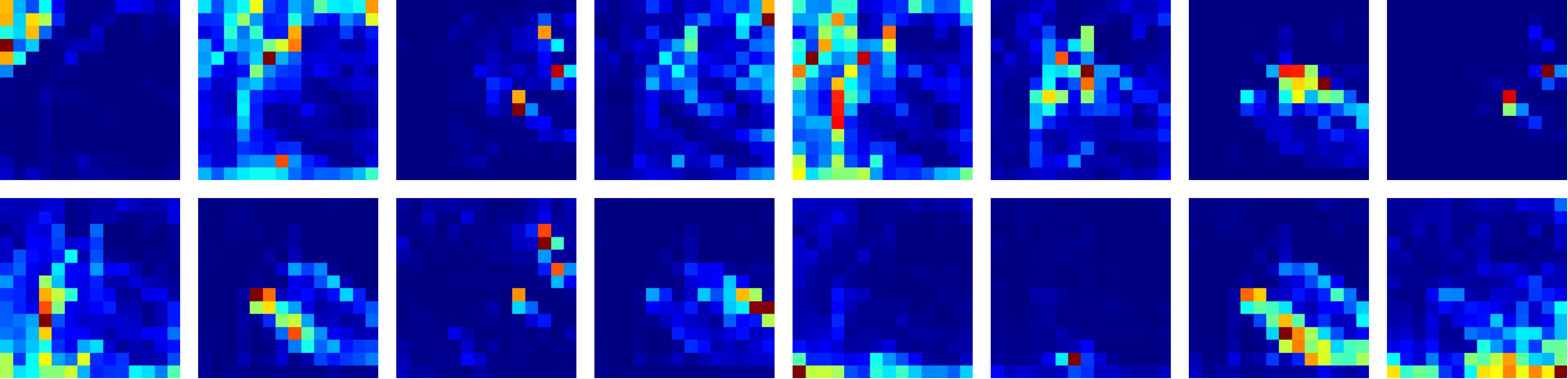}
\centering
\caption{The attention maps ($14\times 14$) of 16 semantic tokens in the second attention layer of STGM.}
\label{app vis c}
\end{subfigure}

\begin{subfigure}{0.45\textwidth}
\centering
\includegraphics[width=1.0\linewidth]{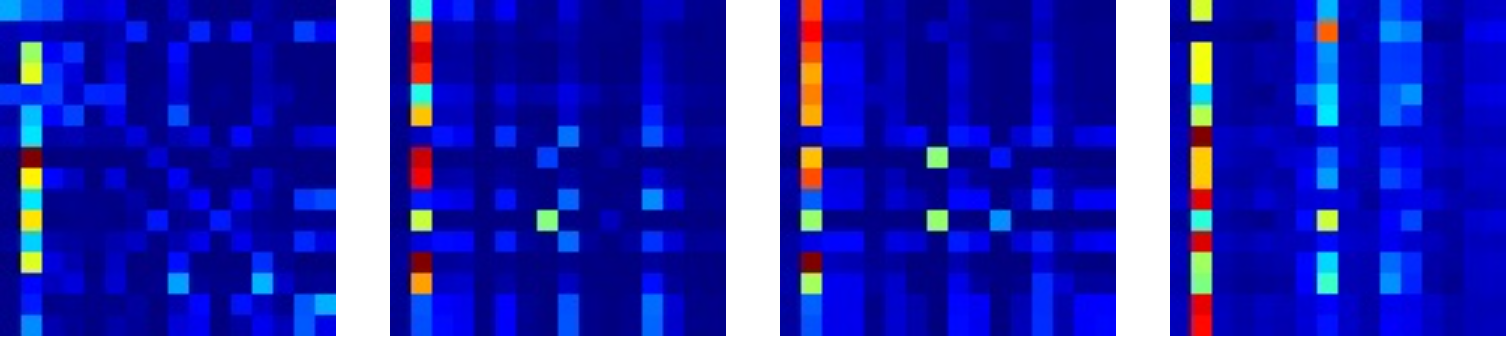}

\caption{Some examples of attention maps ($16\times 16$) of the self-attention layers after STGM.}
\label{app vis b}
\end{subfigure}

\begin{subfigure}{1.\textwidth}
\includegraphics[width=0.9\linewidth]{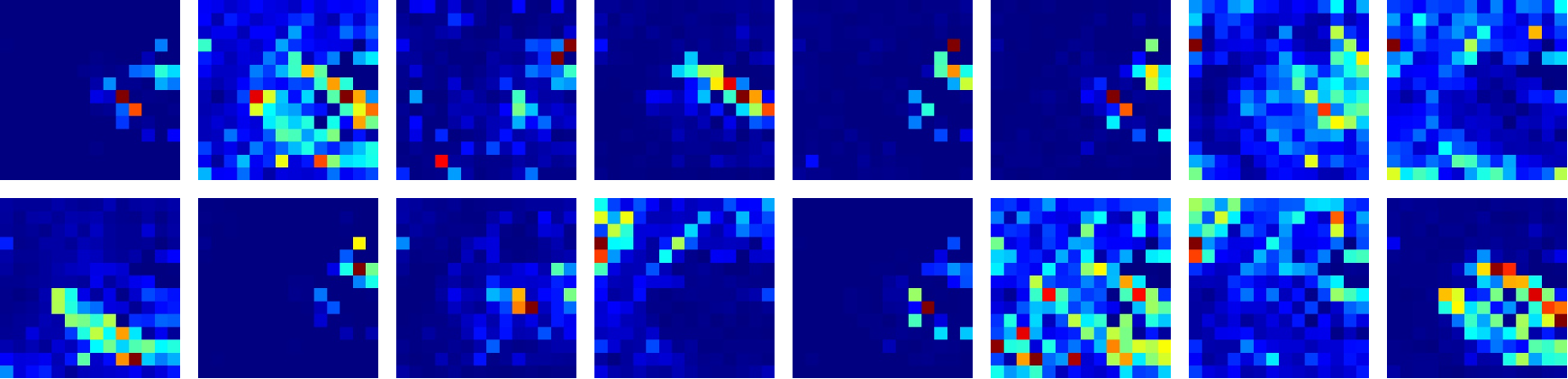}
\centering
\caption{The attention maps ($14\times 14$) of 16 semantic tokens generated by single global initialization.}
\label{app vis d}
\end{subfigure}

\centering
\caption{Additional visualization of attention maps.
}
\label{app vis}
\end{figure*}

We visualize the attention map of the second attention layer in STGM in Figure \ref{app vis a}. The shape of attention map is $N_s\times (N_s + N_i)$, where $N_s=16$, and $N_i=196$. 
The results of the attention computation between semantic tokens $S^1$ (queries) and semantic tokens $S^1$ (keys) are shown in the most left 16 columns, and the rest columns show the computation between semantic tokens $S^1$ and image tokens $X$. The figure shows that the second semantic token highlights the region of semantic tokens, while other semantic tokens highlight the image tokens. Figure \ref{app vis b} visualizes the attention maps in the self-attention layers after STGM. The second semantic token is incorporated by the majority of semantic tokens. These phenomenons illustrate that the second semantic token focuses on more global semantic information, which further verifies our global cluster center initialization can guide the semantic tokens to extract global semantic information. The phenomenons in Figure \ref{app vis a} and Figure \ref{app vis b} nearly emerge in all the images.

Neglecting the most left 16 columns of Figure \ref{app vis a}, we reshape it into 16 $14\times 14$ attention maps like Figure \ref{vis} and show them in Figure \ref{app vis c}. Thanks to the clustering of second attention and global initialization G, we can see that the semantic information is more accurate and meaningful. 

We visualize the attention maps of semantic tokens with single global initialization in Figure \ref{app vis d}. Without spatial initialization, the response regions are more global and similar. In contrast, our semantics of each semantic token are associated with the specific spatial location, which is the basis to allow our method to be applied in local self-attention and downstream tasks. Additionally, our attention maps contain more recognized and diverse semantic information, reflecting the effectiveness of our spatial initialization.


\begin{table}[t]
\small
\begin{center}
\begin{tabular}{cccc}
            \toprule
            Method & Top-1 Acc & FLOPs(G) \\
            \midrule
            EViT~\cite{liang2022evit} & 82.5(-0.8) & 3.9(-41\%)\\
            EViT~\cite{liang2022evit} & 83.0(-0.3) & 4.7(-29\%)\\
            \textbf{STViT(Ours)} & 82.7(-0.6) & 3.7(-44\%)\\
            \textbf{STViT(Ours)} & 83.1(-0.2) & 4.6(-30\%)\\
            \bottomrule
        \end{tabular}
\vspace{-3mm}
\end{center}
\caption{Comparisons with the state-of-the-art token sparsification method EViT on LV-ViT-S.\vspace{-3mm}}
\label{sota2}
\end{table}

\subsection{Additional ablation study}
All the following experiments of STViT and STViT-R are conducted on DeiT-S and Swin-S unless otherwise specified, respectively. 

\paragraph{The position of STGM.}
The effects of different position of STGM are shown in Table \ref{pos of stgm}. Two transformer layers are employed in STGM in all the experiments. We can see that the performance achieves improvement with appropriately moving the STGM towards deep layers due to better features of image tokens.

\paragraph{Positional encoding.} We try to apply positional encoding to our semantic tokens. Table \ref{pos} shows comparisons of different positional encoding methods, including learned positional encoding, conditional positional encoding, and relative positional encoding~\cite{liu2021swin}. All the positional encoding methods do not work on DeiT-S and Swin-T, even though relative positional encoding improves Swin-T by 1.2\%. These experiments demonstrate that the interaction between our semantic tokens depends on high-level semantic information and nearly does not use position relationships.

\begin{table}[t]
\small
\begin{tabular}{c|c|c|c}
\toprule
Method & Backbone & mIoU & FLOPs(G)\\
\midrule
UperNet & Swin-S & 49.3 & 49 \\
UperNet & STViT-R-Swin-S & 48.3 & 34(-31\%) \\
\midrule
UperNet & Swin-B & 49.7 & 87\\
UperNet & STViT-R-Swin-B & 48.9 & 60(-31\%) \\
\bottomrule
\end{tabular}
\caption{Results of semantic segmentation on the ADE20K val set. A multi-scale inference with resolution $[0.5, 0.75, 1.0, 1.25, 1.5, 1.75]\times$ is applied. FLOPs and latency are measured in backbones with resolution $512 \times 512$.}
\vspace{-2mm}
 \label{seg}
\end{table}

\begin{table} [t]
\small
\begin{center}
\small
\begin{tabular}{c|cccccc}
\toprule
Pos. & 3-5 & 4-6 & 5-7 & 7-9 & 8-10 & 10-12 \\
\midrule
Top-1 Acc(\%) & 79.3 & 79.8 & 79.8 & 80.3 & 80.3 & 79.8\\
FLOPs(G) & 1.56 & 1.91 & 2.25 & 2.95 & 3.30 & 4.00\\
\bottomrule
\end{tabular}
\vspace{-5mm}
\end{center}
\caption{Performance evaluation on the different positions of our STGM.}
\label{pos of stgm}
\end{table}

\begin{table}[t]
\small
\begin{tabular}{c|c|c}
\toprule
 & STViT-DeiT-S Acc & STViT-Swin-T Acc \\
\midrule
Learned & 79.6 & 81.5  \\
Conditional & 79.7 & 81.4 \\
Relative & 79.8 & 81.3 \\
No pos. & 79.8 & 81.5 \\
\bottomrule
\end{tabular}
\vspace{-2mm}
\caption{Performance evaluation on different positional encoding methods. \textit{Learned}, \textit{Conditional}, and \textit{Relative} indicate learned positional encoding, conditional positional encoding, and relative positional encoding, respectively.}
\label{pos}
\end{table}
\vspace{-5mm}

\paragraph{Alternative schemes of spatial pooling.}
We use an intra and inter-window spatial pooling in STGM to generate initial cluster centers, which adaptively save meaningful semantic information and avoids overlap between adjacent windows as much as possible. Furthermore, we explore more spatial pooling schemes, including: (i) spatial pooling with large-size kernel and overlap, (ii) multi-scale spatial pooling, and (iii) adaptive spatial pooling. We adopt 25 semantic tokens in these experiments. In (i), the kernel size and overlap are set to 6 and 4, respectively. In (ii), we use two adaptive pooling layers which produce 9 and 16 tokens separately. The results are presented in Table \ref{adaptive spatial pooling} on DeiT-T. We can see that overlap and multiple scales cannot boost the performance, which also demonstrates that discrete semantic tokens with high-level semantic information benefit our method.

\begin{table}[t]
\small
\resizebox{!}{0.55cm}{
\begin{tabular}{c|c|c|c|c}
\toprule
 & Scheme i & Scheme ii & Adaptive spatial pooling & Ours \\
\midrule
Top-1 Acc(\%)  & 71.6 & 71.7 & 71.9 & 72.2 \\
\bottomrule
\end{tabular}}
\vspace{-2mm}
\caption{Alternative schemes of spatial pooling.}
\label{adaptive spatial pooling}
\end{table}


\onecolumn
\subsection{Cluster center recovery by self attention}\label{justification}
We present an analysis showing how cluster centers are recovered through the attention mechanism. Let $K$ be the number of clusters. Let $\N(\mu_i, \sigma^2 I/d), i=1, \ldots, K$ be the $K$ Gaussian distributions, with center $\mu_i \in \R^d$ and covariance matrix $\sigma^2 I/d$. Let $x_{i,j} \in \R^d, j=1, \ldots, n,$ be the $n$ data points independently sampled from $\N(\mu_i, \sigma^2 I /d)$. Given data points $\D = \left\{x_{i,j}, i\in[K], j \in [n]\right\}$, of course without knowing the association of each data point to its underlying Gaussian distribution, our goal is to recover the underlying cluster centers $\mu_i, i \in [K]$. We assume that all the center vectors of Gaussian distributions are well separated, i.e. $\langle \mu_j, \mu_k \rangle \leq \gamma$ if $j \neq k$. For the convenience of study, we assume $|\mu_i| = 1, i \in [K]$.

Let $\mh_i \in \R^d, i \in [K]$ the initialized cluster centers, with all the cluster centers being well normalized. Define $\Delta$ as the gap for any initialized $\mh_i$ to the target cluster centers $\mu_i$ than to other clusters $\mu_j$, i.e. 
\[
    \Delta = \min\limits_{i \in [K]} \min\limits_{j \neq i} \langle \mh_i, \mu_i - \mu_j \rangle
\]
The new cluster centers are estimated through the self-attention mechanism, i.e. 
\[
    \mh_k' = \frac{1}{Z_k}\sum_{i=1}^K \sum_{j=1}^n \exp\left(\lambda\langle \mh_k, x_{i,j}\rangle \right)x_{i,j}
\]
where $\lambda > 0$ is a scaling factor and $Z_i$ is defined as
\[
    Z_k = \sum_{i=1}^K \sum_{j=1}^n \exp\left(\lambda\langle \mh_k, x_{i,j}\rangle \right)
\]
\begin{thm}
With sufficiently large $d$ and $n \gg d$, with a probability $1 - O(K/n^2)$, we have
\[
\frac{\langle \mu_k, \mh_k'\rangle}{|\mh_k'|} \geq 1 - O\left(\frac{\log K + \log d}{d\Delta}\right)
\]
\end{thm}
Define $u_{i,j} = x_{i,j} - \mu_i$. We have
\[
\mh_k' = \frac{1}{Z_k}\sum_{i=1}^K \exp\left(\lambda\langle \mu_i, \mh_k \rangle\right)\left\{\left(\sum_{j=1}^n\exp\left(\lambda\langle \mh_k, u_{i,j} \rangle\right)\right)\mu_i + \sum_{j=1}^n \exp\left(\lambda\langle \mh_k, u_{i,j} \rangle\right)u_{i,j}\right\}
\]
We first bound $\sum_{j=1}^n \exp\left(\lambda \langle \mh_k, u_{i,j}\rangle \right)$. Since $u_{i,j} \sim \N(0, \sigma^2 I /d)$ and $|\mh_k| = 1$, we know that $\langle \mh_k, u_{i,j} \rangle \sim \N(0, \sigma^2/d)$. Hence, with a probability $1 - 2\delta$, we have
\[
\left|\sum_{j=1}^n \exp\left(\lambda \langle \mh_k, u_{i,j}\rangle \right) - n\E_{x\sim\N(0,\sigma^2/d)}\left[\exp(\lambda x)\right]\right| \leq 3\exp\left(\lambda\sigma\sqrt{\frac{2}{d}\log\frac{n}{\delta}}\right) + 2\sqrt{n\E_{x\sim\N(0,\sigma^2/d)}\left[\exp(2\lambda x)\right]\log\frac{2}{\delta}}
\]
Since
\[
\E_{x\sim\N(0,\sigma^2/d)}\left[\exp(\lambda x)\right] = \sqrt{\frac{d}{2\pi\sigma}}\int^{+\infty}_{-\infty}\exp\left(\lambda x - \frac{x^2 d}{2\sigma^2}\right) dx = \exp\left(\frac{\lambda^2\sigma^2}{2d}\right)
\]
we have, with a probability $1 - 2\delta$,
\[
\left|\sum_{j=1}^n \exp\left(\lambda \langle \mh_k, u_{i,j}\rangle \right) - n\exp\left(\frac{\lambda^2\sigma^2}{2d}\right)\right| \leq 3\exp\left(\lambda\sigma\sqrt{\frac{2}{d}\log\frac{n}{\delta}}\right) + 2\sqrt{n\exp\left(\frac{2\lambda^2\sigma^2}{d}\right)\log\frac{2}{\delta}}
\]
With large enough $n$, we have
\[
3\exp\left(\lambda\sigma\sqrt{\frac{2}{d}\log\frac{n}{\delta}}\right) + 2\sqrt{n\exp\left(\frac{2\lambda^2\sigma^2}{d}\right)\log\frac{2}{\delta}} \leq C\sqrt{n}\exp\left(\frac{\lambda^2\sigma^2}{2d}\right)
\]
and therefore
\[
    (1 - \tau)n\exp\left(\frac{\lambda^2\sigma^2}{2d}\right)\leq \sum_{j=1}^n \exp\left(\lambda\langle \mh_k, u_{i,j} \rangle\right) \leq (1 + \tau)n\exp\left(\frac{\lambda^2\sigma^2}{2d}\right)
\]
where
\[
    \tau \leq \frac{C}{\sqrt{n}}
\]
Here $C>0$ is a universal constant. 

We second bound $\sum_{j=1}^n \exp\left(\lambda\langle\mh_k, u_{i,j}\rangle\right) u_{i,j}$. We write each $u_{i,j} = u^{\perp}_{i,j} + u^{\parallel}_{i,j}$, where $u^{\parallel}_{i,j} = \langle u_{i,j}, \mh_k\rangle \mh_k$ and $u_{i,j}^{\perp}$ is a $d-1$ dimensional Gaussian vector. We have
\begin{eqnarray*}
\sum_{j=1}^n \exp\left(\lambda\langle\mh_k, u_{i,j}\rangle\right) u_{i,j} = \left(\sum_{j=1}^n \exp\left(\lambda\langle \mh_k, u_{i,j}\rangle\right)\langle \mh_k, u_{i,j}\rangle\right)\mh_k + \sum_{j=1}^n u^{\perp}_{i,j} 
\end{eqnarray*}
Since $u^{\perp}_{i,j}\sim\N\left(0, \sigma^2 I_{d-1}/d\right)$, we have $\sum_{j=1}^n u^{\perp}_{i,j} \sim \N\left(0, n\sigma^2 I_{d-1}/d\right)$. Using the concentration of $\chi^2_{d-1}$ distribution, we have, with a probability $1 - \delta$
\[
\left|\sum_{j=1}^n u^{\perp}_{i,j}\right|^2 \leq \frac{n\sigma^2}{d}\left(d-1 + 2\sqrt{(d-1)\log\frac{1}{\delta}} + 2\log\frac{1}{\delta}\right) \leq n\sigma^2\left(1 + 3\sqrt{\frac{\log(1/\delta)}{d}}\right)
\]
To bound $\sum_{j=1}^n \exp\left(\lambda\langle \mh_k, u_{i,j}\rangle\right)\langle \mh_k, u_{i,j}\rangle$, following the same procedure, we have, with a probability $1 - 2\delta$
\begin{eqnarray*}
\lefteqn{\left|\sum_{j=1}^n \exp\left(\lambda\langle \mh_k, u_{i,j}\rangle\right)\langle \mh_k, u_{i,j}\rangle - n\E_{x\sim\N(0,\sigma^2/d)}\left[\exp(\lambda x)x\right]\right|} \\
& \leq & 3\exp\left(\lambda\sigma\sqrt{\frac{2}{d}\log\frac{n}{\delta}}\right)\sigma\sqrt{\frac{2}{d}\log\frac{n}{\delta}} + \sqrt{n\E_{x\sim\N(0,\sigma^2/d)}\left[\exp(2\lambda x) x^2\right]\frac{2}{\delta}}
\end{eqnarray*}
Since
\[
\E_{x\sim\N(0,\sigma^2/d)}\left[\exp(\lambda x)x\right] = \sqrt{\frac{d}{2\pi\sigma^2}}\int\exp\left(\lambda x - \frac{x^2 d}{2\sigma^2}\right) x dx = \frac{\lambda\sigma^2}{d}\exp\left(\frac{\lambda^2\sigma^2}{2d}\right)
\]
and
\begin{eqnarray*}
\lefteqn{\E_{x\sim\N(0,\sigma^2/d)}\left[\exp(2\lambda x)x^2\right]} \\
& = & \sqrt{\frac{d}{2\pi\sigma^2}}\int\exp\left(2\lambda x - \frac{x^2 d}{2\sigma^2}\right) x^2 dx \\
& = & \sqrt{\frac{d}{2\pi\sigma^2}}\exp\left(\frac{2\lambda^2\sigma^2}{d}\right)\int\exp\left(\frac{ d}{2\sigma^2}\left[x - \frac{\lambda\sigma^2}{d}\right]^2\right) \left(\left[x - \frac{\lambda\sigma^2}{d}\right]^2 + 2\frac{\lambda\sigma^2}{d}\left[x - \frac{\lambda\sigma^2}{d}\right] + \frac{\lambda^2\sigma^4}{d^2}\right) dx \\
& = & \exp\left(\frac{2\lambda^2\sigma^2}{d}\right)\left(\frac{\lambda^2\sigma^4}{d^2} + \left(\frac{2\sigma^2}{d}\right)^{3/2}\E_{x\sim\N(0,1)}\left[x^2\right]\right) \\
& = & \exp\left(\frac{2\lambda^2\sigma^2}{d}\right)\left(\frac{\lambda^2\sigma^4}{d^2} + 2\left(\frac{2\sigma^2}{d}\right)^{3/2}\right) \leq \frac{2\lambda^2\sigma^4}{d^2}\exp\left(\frac{2\lambda^2\sigma^2}{d}\right)
\end{eqnarray*}
We thus have, with a probability $1 - 2\delta$, 
\begin{eqnarray*}
\lefteqn{\left|\sum_{j=1}^n \exp\left(\lambda\langle \mh_k, u_{i,j}\rangle\right)\langle \mh_k, u_{i,j}\rangle - \frac{n\lambda\sigma^2}{d}\exp\left(\frac{\lambda^2\sigma^2}{2d}\right)\right|} \\
& \leq & 3\exp\left(\lambda\sigma\sqrt{\frac{2}{d}\log\frac{n}{\delta}}\right)\sigma\sqrt{\frac{2}{d}\log\frac{n}{\delta}} + \sqrt{\frac{4n\lambda^2\sigma^4}{d^2}\exp\left(\frac{2\lambda^2\sigma^2}{d}\right)\log\frac{2}{\delta}}
\end{eqnarray*}
When $n$ is sufficiently large, we have, with a probability $1 - 2\delta$
\[
\left(1 - \tau\right)\frac{n\lambda\sigma^2}{d}\exp\left(\frac{\lambda^2\sigma^2}{2d}\right) \leq \sum_{j=1}^n \exp\left(\lambda\langle \mh_k, u_{i,j}\rangle\right)\langle \mh_k, u_{i,j}\rangle \leq \left(1 + \tau\right)\frac{n\lambda\sigma^2}{d}\exp\left(\frac{\lambda^2\sigma^2}{2d}\right)
\]
where $\tau \leq C/\sqrt{n}$. By putting them together, with a probability $1 - 4\delta$, we have
\[
\sum_{j=1}^n \exp\left(\lambda \langle \mh_k, u_{i,j}\rangle\right)u_{i,j} = n(1+\beta)\frac{n\lambda\sigma^2}{d}\exp\left(\frac{\lambda^2\sigma^2}{2d}\right)\mh_k + n\nu_i
\]
with $\beta \in [1-\tau, 1+\tau]$ and 
\[
    |\nu_i| \leq \frac{2\sigma}{\sqrt{n}}
\]
Finally, we have, with a probability $1 - 4K\delta$
\[
\mu'_k = \frac{n}{Z_k}\sum_{i=1}^K\exp\left(\lambda\langle \mu_i, \mh_k\rangle\right)\left((1 + \alpha_i)\exp\left(\frac{\lambda^2\sigma^2}{2d^2}\right)\mu_i + (1+\beta_i)\frac{\lambda\sigma^2}{d}\exp\left(\frac{\lambda^2\sigma^2}{2d}\right)\mh_k + \nu\right)
\]
Using the same analysis, we have, with a probability $1 - 4K\delta$,
\[
    \frac{Z_k}{n} = \exp\left(\frac{\lambda^2\sigma^2}{2d^2}\right)\sum_{i=1}^K\exp\left(\lambda\langle\mu_i, \mh_k\rangle\right)(1 + \alpha_i) 
\]
Now, we can bound $|\mh_k' - \mu_k|$. We have
\begin{eqnarray*}
\lefteqn{|\mu_k - \mh_k'|} \\
& \leq & \left|\frac{n}{Z_k}\exp\left(\lambda\langle \mu_k, \mh_k\rangle + \frac{\lambda^2\sigma^2}{2d^2}\right)(1+\alpha_k) - 1\right| + \frac{n}{Z_k}\sum_{i\neq k}\exp\left(\lambda\langle \mu_i, \mh_k\rangle + \frac{\lambda^2\sigma^2}{2d^2}\right)(1+\alpha_i) \\
&  & + \frac{n\lambda\sigma^2|\mu_k - \mh_k|}{Z_k d}\sum_{i=1}^K\exp\left(\lambda\langle \mu_i, \mh_k\rangle + \frac{\lambda^2\sigma^2}{2d^2}\right)(1+\beta_i) + \frac{n}{Z_k}\sum_{i=1}^K\exp\left(\lambda\langle \mu_i, \mh_k\rangle\right)\nu_i
\end{eqnarray*}
To further develop the bound for $|\mu_k - \mh_k'|$, we have
\[
Z_k \geq n\exp\left(\frac{\lambda^2\sigma^2}{2d^2} + \lambda\langle \mu_k, \mh_k\rangle\right)\left(1 - \frac{C}{\sqrt{n}}\right)
\]
and
\[
Z_k \leq  n\exp\left(\frac{\lambda^2\sigma^2}{2d^2} + \lambda\langle \mu_k, \mh_k\rangle\right)\left(1 + \frac{C}{\sqrt{n}}\right)\left(1 + (K-1)\exp\left(-\lambda\Delta\right)\right)
\]
We thus have
\begin{eqnarray*}
\lefteqn{|\mu_k - \mh_k'|} \\
& \leq & \left|\frac{\exp\left(\lambda\Delta\right)}{\exp\left(\lambda\Delta\right) + K - 1}\left(1 - \frac{2C}{\sqrt{n}}\right)^2 - 1\right| + \left(1 + \frac{2C}{\sqrt{n}}\right)^2\frac{K-1}{\exp(\lambda\Delta)}\left(1 + \frac{\lambda\sigma^2}{d}|\mu_k - \mh_k|\right) \\
&  & + \frac{\lambda\sigma^2}{d}|\mu_k - \mh_k| + \left(1 + \frac{2C}{\sqrt{n}}\right)\frac{\sigma}{\sqrt{n}}
\end{eqnarray*}
By choosing $\lambda = (\log d + \log K)/\Delta$, and by assuming $n$ is significantly larger than $d$, we have
\[
|\mu_k - \mh_k'| \leq O\left(\frac{\log d + \log K}{d\Delta}\right)
\]
implying that
\[
    \frac{\langle \mu_k, \mh'_k \rangle}{|\mh_k'|} \geq 1 - O\left(\frac{\log d + \log K}{d\Delta}\right)
\]

\end{document}